\documentclass[10pt,a4paper,twoside]{amsart}

\usepackage[utf8]{inputenc} 
\usepackage[UKenglish]{babel} 
\usepackage[T1]{fontenc} 
\usepackage{xcolor} 
\usepackage{microtype, fancyhdr, lmodern}
\usepackage[subtle]{savetrees}
\usepackage{soul}
\usepackage[a4paper, hmarginratio=1:1]{geometry}

\usepackage{amsfonts, amsmath, amsthm, amssymb, mathrsfs, amscd}	
\usepackage{blkarray}
\numberwithin{equation}{section}	
\usepackage{faktor}
\usepackage{siunitx}
\allowdisplaybreaks

\usepackage{url, enumitem}
\usepackage{algorithm}
\usepackage{algorithmicx}
\usepackage[noend]{algpseudocode}

\usepackage{longtable}
\usepackage{tabularray}
\usepackage{booktabs, array}
\newcolumntype{C}[1]{>{\centering\arraybackslash}p{#1}}
\usepackage{multirow,bigdelim, caption}
\usepackage{tikz}
\usetikzlibrary{shapes,arrows,positioning, arrows.meta}
\usepackage{graphicx}
\usepackage{float}
\usepackage{subcaption}
\graphicspath{{Figures/}}
\usepackage{tabularray}

\newcommand\Nn{\mathbb{N}}

\newcommand\Rr{\mathbb{R}}

\theoremstyle{plain}

\theoremstyle{remark}

\theoremstyle{definition}

\makeatletter
\newenvironment{breakablealgorithm}
  {
   \begin{center}
     \refstepcounter{algorithm}
     \hrule height.8pt depth0pt \kern2pt
     \renewcommand{\caption}[2][\relax]{
       {\raggedright\textbf{\fname@algorithm~\thealgorithm} ##2\par}%
       \ifx\relax##1\relax 
         \addcontentsline{loa}{algorithm}{\protect\numberline{\thealgorithm}##2}%
       \else 
         \addcontentsline{loa}{algorithm}{\protect\numberline{\thealgorithm}##1}%
       \fi
       \kern2pt\hrule\kern2pt
     }
  }{
     \kern2pt\hrule\relax
   \end{center}
  }
\makeatother

\usepackage[bookmarks,bookmarksnumbered]{hyperref}
\hypersetup{colorlinks = true,linkcolor = blue}

\begin{document}

\begin{abstract}
Federated learning is becoming an increasingly viable and accepted strategy for building machine learning models in critical privacy-preserving scenarios such as clinical settings. Often, the data involved is not limited to clinical data but also includes additional omics features (e.g. proteomics). Consequently, data is distributed not only across hospitals but also across omics centers, which are labs capable of generating such additional features from biosamples. This scenario leads to a hybrid setting where data is scattered both in terms of samples and features. In this hybrid setting, we present an efficient reformulation of the Kernel Regularized Least Squares algorithm, introduce two variants and validate them using well-established datasets. Lastly, we discuss security measures to defend against possible attacks. 
\end{abstract}


\title{A Hybrid Federated Kernel Regularized Least Squares Algorithm}

\author{Celeste Damiani}
\address{Computational and Chemical Biology, Fondazione Istituto Italiano di Tecnologia, Morego 30, Genoa, 16163, Italy}
\email{celeste.damiani@iit.it}

\author{Yulia Rodina}
\address{Computational and Chemical Biology, Fondazione Istituto Italiano di Tecnologia, Morego 30, Genoa, 16163, Italy}
\email{yulia.rodina@iit.it}

\author{Sergio Decherchi}
\address{Data Science and Computation Facility, Fondazione Istituto Italiano di Tecnologia, Morego 30, Genoa, 16163, Italy}
\email{sergio.decherchi@iit.it}

\date{\today}

\maketitle

\section{Introduction}
Nowadays, it is getting more and more appealing to jointly apply machine learning algorithms to clinical and omics data. By omics data, we mean data that is generated by specialized labs in the field of proteomics, genomics, transcriptomics and lipidomics, to name some examples.
It is often the case that clinical and omics data cannot be shared among hospitals, research centers and other parties that might be involved in healthcare research, because of issues with data privacy, data security, data access rights and heterogeneity of data~\cite{zhou_ppml-omics_2022}.  
Federated Learning (FL) is a machine learning approach that allows overcoming this issue and collaboratively train a model using the union of data coming from different ``data islands'', without merging local datasets or exposing any data information. This approach protects patients' data privacy while enabling healthcare organizations to collaborate and train models~\cite{Babar:2024}.
Federated learning is usually divided into \emph{horizontal FL}, \emph{vertical FL} and \emph{hybrid FL}, according to the partitioning of data. In the horizontal setting, each participant to the federation has the data of a different set of subjects, and the data of every client have the same features; analogously, in the vertical setting the same overall set of samples is shared among the different clients owning different features for them~\cite{li_survey_2021}. However, in a collaborative research platform like the one we are interested in building, each client might have access to only certain subjects and certain features: this is what is called a hybrid setting. While this scenario is extremely common in practical settings, at the moment of writing few methods exist for hybrid FL~\cite{overman_primal-dual_2024}.
In particular, we are thinking of several hospital-clients, each with their own set of patients, for which several research institutes or omics centers have different features.
The state-of-the-art of developing new models or rewriting of current models adapted to a federated setting is evolving quickly~\cite{liu2023recent}. References in this text are by no means exhaustive, but aim to give the reader some landmarks to start navigating this quickly evolving field.

One of the first and most popular algorithms to train neural networks in a federated way is the stochastic gradient descent, see for instance~\cite{mcmahan_communication-efficient_2017, wang_federated_2020, Huimin:2023}. Decision trees have also been widely used, especially for vertical or hybrid settings. In particular, gradient boosting decision trees proved to have a good performance in classification and regression tasks~\cite{li_practical_2020, wu_privacy_2020, cheng_secureboost_2021}.
Besides neural networks and decision trees, also linear models~\cite{Cellamare:2022, Ghosh:2022} and in particular logistic regression \cite{Deist:2020, Vaid:2021, He:2022}, linear regression \cite{Mandal:2019}, SVMs \cite{yu_privacy-preserving_2006, mangasarian_privacy-preserving_2008} and more general kernel methods \cite{polato_privacy-preserving_2021, hannemann_privacy-preserving_2023} have been studied in a federated context.

Many other choices need to be made when setting up a FL platform. We refer to~\cite{li_survey_2021} for an overarching taxonomy of the topic articulated around the main challenges of a federated architecture, such as data partitioning, choice of models, communication architecture, scale of federation, privacy-preserving mechanisms, management of non-iid data.
We also refer to~\cite{liu2023recent} for an alternative taxonomy, centered around the different approaches to setting up a pipeline in four different cases: aggregation optimization, heterogeneous federated learning, secure federated learning and fair federated learning.

Federated kernel methods have been studied both in a horizontal~\cite{hannemann_privacy-preserving_2023} and a vertical~\cite{polato_privacy-preserving_2021} setting. Techniques range from leveraging properties of dot products and adding some masking as in the above cited papers, to adding noise to the data as in~\cite{Rubinstein_Bartlett_Huang_Taft_2012}. A strategy can be to share only Gram matrices, and representing kernel values to preserve the raw data~\cite{Kilho_2024}.

In this work we introduce two learning procedures we call \emph{FedCG}: a non-iterative, very fast one based on a mild assumption which we verify practically to ensure security, and one that is unconditionally secure and iterative. In both the cases, we ensure that the solution is identical to the non-federated version thanks to the convexity of the cost function. To the best of our knowledge, in the hybrid FL setting this has only been achieved in~\cite{overman_primal-dual_2024}, using Fenchel Duality to solve convex problems. That approach is substantially different from ours, as it is based on the primal-dual algorithm whereas we maintain only a dual representation.
We base our algorithms on two key observations. The first one is that an RBF kernel matrix can be expressed as a Hadamard product of kernel matrices, each associated to a specific feature. The second one is that one can adapt the Nystr{\"o}m approximation, by replacing randomly selected landmarks with fully random landmarks~\cite{Gastaldo-Decherchi:2016}. These key observations easily extend to other kernel-based methods rendering our algorithm more like a general approach. To solve the implied optimization problems we take advantage of the conjugate gradient method. 

The paper is organized as follows: after providing a comprehensive formulation of the introduced FedCG algorithms, 
in Subsection~\ref{SS:fedCGintro} we discuss how we can ensure the security of our algorithm, at the price of some communication overhead, see Subsection~\ref{SS:securefedCG}. In Subsection~\ref{SS:Performance} we experimentally show the efficacy of our approach in achieving convergence within this hybrid federated setting, aligning with the performance benchmarks set by the centralized version of the algorithm. Finally, in Subsection~\ref{SS:EDM} we will discuss an extra security, in the light of available solutions to the Euclidean Distance Matrix Completion Problem~\cite{Liberti:survey2014}.

\section{Methods}

In this section we present privacy-preserving techniques for solving the optimization problem underlying the Kernel Regularized Least Squares method (KRLS) in a hybrid FL framework. Federated computation of kernel matrices has been mostly focused on SVMs within the horizontal FL context. We refer to~\cite{polato_privacy-preserving_2021} for an overview of recent works, and for a novel technique of carrying on such calculations in a vertical FL setting.

Let us consider a training set $\{ (x_1, y_1), \ldots, (x_n, y_n)\}$ where each $x_i$ is a $d$-dimensional column vector. We denote by $X$ the $n$-by-$d$ matrix whose $i$-th row is $x_i^T$, and by $y = [y_1, y_2, \ldots, y_n]^T$ a column vector of outputs or labels.
The problem KRLS solves consists in finding a non linear function $f$ that minimizes the standard least squares cost function, with the introduction of a Tikhonov  regularization operator whose weight is ruled by a positive parameter $\lambda$. This operator, for proper values of the $\lambda$ hyperparameter ensures generalization. KRLS can be used for both regression and classification problems.
The optimization problem of KRLS is:
\begin{equation}
    \label{E:RLS}
    \frac{1}{n} \sum_{i=1}^{n} (y_i - f(x_i))^2 + \lambda ||f||_{\mathcal{H}}^2
\end{equation}

where $f$ is a function such that $f(x)$ is an effective estimate of the output $y$ when a new input $x$ is given, and where the minimization takes place on a Reproducing Kernel Hilbert Space of continuous functions~$\mathcal{H}$. For references on this classical problem see for instance \cite{Poggio-Girosi:1990, evgeniou_regularization_2000, cucker_best_2002, de_vito_model_2005}.

\begin{figure}
    \centering
    \begin{subfigure}[t]{.45\linewidth}
        \centering
        \includegraphics[width=\linewidth]{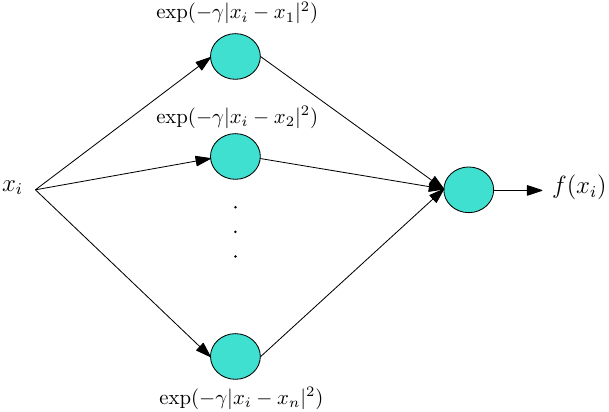}
        \caption{Representation of KRLS with a Gaussian kernel as a neural network, where for each sample $x_i$, we have $n$ equations, that we can see as ``neurons'', corresponding to the entries $K_{i, j} = \exp(-\gamma ||x_i-x_j||^2)$, $j \in \{1, \ldots, n\}$, of the kernel matrix~$K$. }
        \label{F:KRLS}
    \end{subfigure}
    \hfill
    \begin{subfigure}[t]{.45\linewidth}
        \centering
        \includegraphics[width=\linewidth]{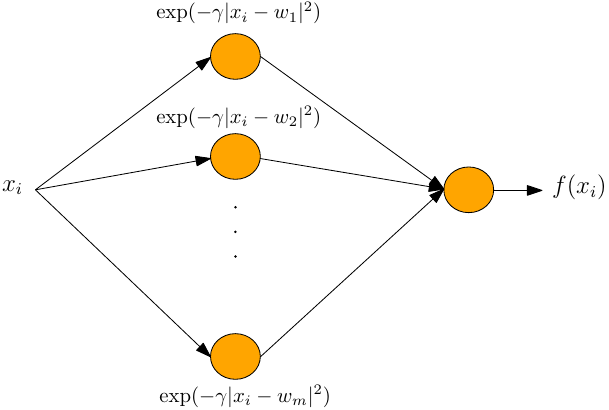}
        \caption{In the RRLS, we chose random centers $w_1, \ldots, w_m \in \Rr^d$, with $m \leq n$, distributed with an arbitrary distribution. This allows to both reduce the computational load, the required memory and is privacy preserving.}
        \label{F:RNKRLS}
    \end{subfigure}
    \caption{Visual comparison between KRLS and RRLS.}
    \label{fig:main}
\end{figure}

For simplicity in the following, we can fold the $\frac{1}{n}$ term into the Tikhonov regularization parameter~$\lambda$ and thus rewrite the cost function as:
\begin{equation}
    \label{E:RLS2}
    \sum_{i=1}^{n} (y_i - f(x_i))^2 + 
    \lambda||f||_{\mathcal{H}}^2
\end{equation}
Remark that these changes do not change the minimizer. 
It can be proved that a solution $f$ to \eqref{E:RLS2} exists, is unique, and using the representer theorem, it can be written as:
\begin{equation}
\label{E:representer}
f(\cdot) = \sum_{j=1}^{n} \alpha_j k(x_j, \cdot)
\end{equation}
where $\alpha_j \in \mathbb {R}$ for $1 \leq j \leq n$ and $k$ is a positive semidefinite kernel function. We define the kernel matrix $K$ such that $K_{i,j} = k(x_i, x_j)$, 
and by abuse of notation, we allow the kernel function to take multiple data points and produce a matrix, \textit{i.e.}, $k(X, X) = K$.  In Figure \ref{F:KRLS} we show a graphical representation of KRLS as a single hidden layer neural network.
Because of the representer theorem Equation~\eqref{E:RLS2} can be rewritten as:
\begin{equation}
    \label{E:RLSker}
     ||y - K\alpha||_2^2 + \lambda \alpha^T K \alpha
\end{equation}
where $\alpha$ is the optimal $n$-dimensional vector of coefficients $(\alpha_1, \ldots, \alpha_n) \in \mathbb{R}^n$.
By setting the gradient with respect to $\alpha$ to zero (and carrying out some further rewriting), one can show that to obtain a solution $\alpha$ for a fixed value of the parameter $\lambda$, one needs to solve the 
following linear system of equations:
\begin{equation}
    \label{E:KRLS}
    (K + \lambda I) \alpha = y 
\end{equation}
and so, the problem admits a closed-form solution: 
\begin{equation}
    \label{E:solution}
    \alpha = (K + \lambda I)^{-1} y
\end{equation}

For more details and solving techniques see, for instance, \cite{Hainmueller_Hazlett_2014} and its references.
While effective and exact, in that KRLS has a convex loss function as long as a Mercer kernel is used, memory requirements make the KRLS method (and more generally kernel methods) unfeasible when dealing with large training datasets, see~\cite{Smola:2000, Williams:2000, Drineas:2005, caponnetto:2007, rudi_less_2015} for works exploring a variety of strategies to overcome this issue. 
Among the strategies appearing in the literature to deal with the practical unfeasability to solve KRLS, we find Nystr{\"o}m approaches, that replace the complete kernel matrix, with a smaller matrix obtained by (column) subsampling~\cite{Smola:2000, Williams:2000}. In particular, we adopt a version of KRLS using such a reduced matrix, leveraging the fact that the subsampling level controls both regularization
and computations complexity~\cite{rudi_less_2015}. 
The principle of these approaches is to replace the $f$ representation with:
\begin{equation}
    \label{E:Hm}
      f_m \in \mathcal{H}_m\:,\:  f_m = \sum_{j=1}^{m} \alpha_j k( \tilde{x}_j, \cdot), \alpha_1, \ldots, \alpha_m \in \mathbb{R} 
\end{equation}
where $m \leq n$ and $\{\tilde{x}_1, \ldots, \tilde{x}_m\}$ is the selected subset of the points in the training set. From now on we will refer to these points as \emph{landmarks}. In this setting we cannot appeal to a representer theorem, but based on the interpretation as a neural network~\cite{Husmeier1999} then the solution $f_m$ still represents a single hidden layer neural network.
The corresponding minimization problem can now be written as:
\begin{equation}
    \label{E:NKRLS}
     ||y - K_m\alpha||_2^2 + \lambda ||\alpha||_2^2
\end{equation}
By setting again the gradient to 0 we get:
\begin{equation}
    \label{E:NKRLSsystem}
    (K_m^TK_m + \lambda I) \alpha = K_m^Ty
\end{equation}
 where $K_m$ is an $n \times m$ matrix with entries $(K_m)_{i, j} = K(x_i, \tilde{x}_j)$, for $i \in \{1, \ldots, n\}$ and $j \in \{1, \ldots, m\}$~\cite{rudi_less_2015}.
The problem can be solved via matrix pseudo-inversion (we use the $^\dagger$ symbol):
\begin{equation}
    \label{E:NysSolution}
    \alpha = (K_{m}^TK_m + \lambda K_m)^{\dagger}K_m^T y
\end{equation}
This problem is extremely advantageous both in terms of memory and solution time, as it scales with the number of landmarks and not with the training set size. From now on we will assume that the kernel matrix is multiplicatively separable with respect to features; for instance the universal RBF kernel satisfies this property as $k(x_i,x_j)=\Pi_k^m k(x^k_i,x^k_j)$  where $x \in \mathbb{R}^m$. This is the problem we aim at federating in this manuscript. 

\begin{figure}
    \centering
    \includegraphics[width=0.3\linewidth]{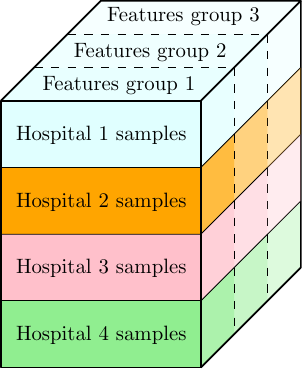}
    \caption{In the hybrid federated setting samples are distributed among several hospitals, and features for said samples are hosted by other entities, such as omics centers. }
    \label{F:fedesetting}
\end{figure}

In our applicative scenario we have one research institute which acts as \emph{federator} (often referred to as server) of several hospitals which have different batches of clinical data of patients, whose features are provided by some omics centers, see Figure~\ref{F:fedesetting}. That is we have fully distributed data both horizontally and vertically.
To solve the issue of entity resolution, we imagine to have a randomly generated patient identifier already shared among the clients (hospitals and omics centers). This is realistic and happens routinely as omics centers produce omics data (e.g. genomics, transcriptomics, \ldots) from biological samples pre-shared by hospitals. As biological samples are physically shipped to omics centers there is not a real problem of identity resolution.

A first difficulty which arises in federating such variant of KRLS is that each hospital-client should have access to the landmarks to build their portion of kernel matrix; however landmarks are training points themselves which cannot be shared among clients. Also, transferring partial kernel matrices (and hence the distance matrix) from clients to an hypothetical server would create a privacy issue as, for instance in the case of the RBF kernel, these could be used to reconstruct the full Euclidean Distance Matrix of the data points, and consequently, the data points configuration (up to roto-translation)\cite{dokmanic_euclidean_2015}. To address this issue we choose to use random points extracted from a distribution (ideally comparable to the one of the data points) as ``Nystr{\"o}m-like'' landmarks instead of subsampling actual points of the dataset. This approach leads to a variant of the KRLS method with Nystr{\"o}m points that we call \emph{Random Nystr{\"o}m KRLS} method (RRLS). See Figure~\ref{F:RNKRLS} for a graphical neural network representation of random Nystr{\"o}m methods. The soundness of applying this method of randomization in the training process has already been studied in the more general context of learning frameworks using similarity functions, such as kernel-based methods and random neural networks~\cite{Gastaldo-Decherchi:2016}. Then the system one needs to solve can still be written as \eqref{E:NKRLSsystem}, except that the points $\tilde{x_j}$ for $\{1, \ldots, m\}$ are not sampled from the training set but can be sampled from another random distribution. We will discuss several methods for sampling in the following. In this paper we employ such learning scheme~\cite{Gastaldo-Decherchi:2016} and provide hybrid federated algorithmic protocols to support the training and ensure privacy.
It is worth stressing that in a federated perspective, this approach addresses the problem that the kernel matrix computations at training time (either whole or subsampled à la Nystr{\"o}m) requires the reference set of landmarks to be shared among the clients. When, instead of using actual training samples as in KRLS or Nystr{\"o}m KRLS,  we generate new samples that do not have any privacy concern, we avoid clients having to share data points (see Subsection~\ref{SS:security} a  discussion on this topic). Note that these newly generated samples are only needed to support the training and don't constitute a new training set. 
\\
To solve iteratively the RRLS problem here formulated, we use the Conjugate Gradient (CG) algorithm, a classical method that can be easily rewritten in a federated way. We change the order of operations in CG to accommodate for the federated setting, insuring 
convergence to the exact solution while adhering to privacy constraints. We  leverage the fact that the kernel matrices that we use can be written as Hadamard products of submatrices: in fact we adapt to the hybrid setting by partitioning the samples matrix in submatrices both vertically and horizontally.

In Section~\ref{SS:securefedCG}, we choose an RBF (Gaussian) kernel matrix, which will contain the distances between the matrix of samples $X$ of dimension $n \times d$, and a random matrix (of random landmarks) $W$ of dimension $m \times d$, with $m \leq n$. In particular, we carried out tests with a matrix $W$ generated with the three following methods:
\begin{description}
    \item[$\mathcal{P}$:] landmarks sampled uniformly at random without replacement from the training set (original Nystr{\"o}m method).
    \item[$\mathcal{U}$:] landmarks randomly generated from the uniform distribution between $0$ and~$1$.
    \item[$\mathcal{N}$:] landmarks randomly generated from a normal distribution with the same center and standard deviation of the training set.
\end{description}
Note that as our data is always normalized in $[0,1]$, hence even in the case of the normal distribution the support of the data and the distribution is the same.

\subsection{A hybrid federated CG optimizer for the RRLS problem.}
\label{SS:fedCGintro}
The classical CG is an iterative optimization technique used to solve systems of linear equations whose matrix is symmetric and positive definite. It was originally introduced in the 50s by Hestenes and Stiefel as a direct method, but it became popular for its properties as an iterative method. 
Given a linear system where $A$ is a symmetric positive definite matrix, the CG method aims to find the solution $x$ that minimizes the residual $r_k = b - A_k$ in a least-squares sense. A key property of the CG method is that the search directions (vector $p$) along which the iterations move are $A$-orthogonal (or conjugate with respect to the matrix $A$), making the search of the solution space more efficient with respect to using the conventional gradient descent procedure. Moreover, the solution is unique, because of the symmetric positive definite condition, and the only operation involving $A$ is multiplication of $A$ by a vector. So if $A$ is large and sparse, the method is computationally efficient. Finally, the number of iterations needed by CG to converge is equal to the number of distinct eigenvalues of $A$, and thus at most equal to the number of samples. For more details about this classical method we refer to~\cite{Shewchuk:painlessCG}.

Such an algorithm needs some degree of adaptation in order to cope with the hybrid federation, so that the matrix $A$ can be distributed.  We describe here two possible solutions to exploit and adapt CG for our aims.

 \subsection{A naive federalized kernel least square learning algorithm}
 \label{SS:naive}
 The first algorithm we propose is a very fast approach to solving RRLS which relies on a mild hypothesis for assuring privacy and for these reasons we call it naive.
 
Given a sample set distributed horizontally among $m \in \Nn$ hospitals with features distributed vertically among a certain number of omics centers (see Figure~\ref{F:fedesetting}), and a Gram matrix $K$ representing a Mercer kernel function $k$, we can rewrite $K$ as follows:
\[
    K = \bigodot_{j \in \textrm{Omics}}
    \begin{pmatrix} 
    K_1^j \\
    K_2^j \\
    \vdots \\
    K_m^j
    \end{pmatrix}
    \]
where we concatenate partial kernel matrices and the operator $\bigodot$ is a Hadamard matrix product. This decomposition is possible thanks to the multiplicative feature-wise separability hypothesis on the kernel.
For every client to be able to compute their block, random landmarks need to be shared. This can be achieved by pre-sharing a common random seeding mechanism among clients for the random generator. Then, applying this rewriting, it is enough that all centers send their $K_i^j$ submatrix to the server, for the server to solve the problem with the CG method.

\begin{figure}[htb]
    \centering
    \includegraphics[width=.8\linewidth]{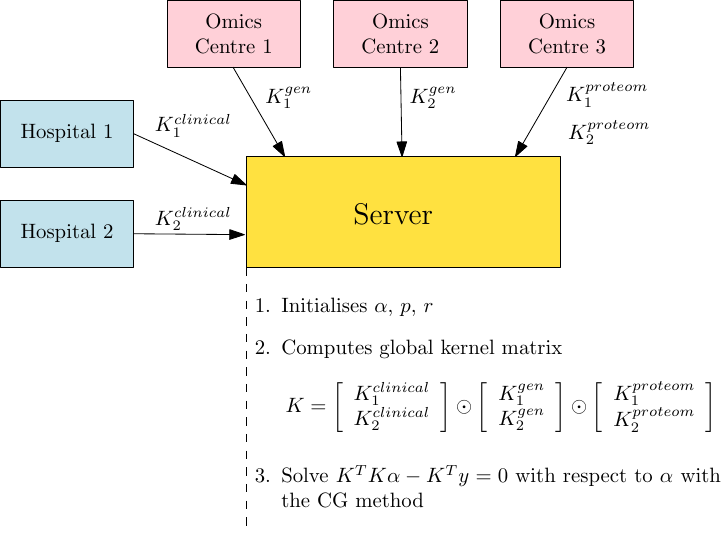}
    \caption{In this example we assume that we are working with patients distributed over two hospitals, who share an array of clinical feature, and three omics centers. The first two provide genomics data (in particular Omics Center 1 provides the genomics features belonging to patients from Hospital 1 and Omics Center 2 the genomics for Hospital 2, while Omics Center 3 provides proteomics features for all the hospitals. Of course, there might be missing data as not all features will have been calculated for all patients.}
    \label{F:naive}
\end{figure}

To solve the problem also the labels are needed.
This is a typical issue in hybrid FL: while in horizontal FL the clients generally do not need to share their labels, in vertical FL the labels need to be either stored by the server or in a designed client~\cite{Chen_2024, Liu_2022}, in hybrid
FL systems, one needs to deal with both types of clients. In several real-world scenarios one can assume that the federator is a center which has a pre-shared access to all the labels. This is very likely to happen when the federator is an omics center. Hence in this case there is no need to protect the labels communication. However, in the proposed methods we address the not-pre-shared labels scenario by protecting communications with synchronized random noise; nevertheless also much more robust plain symmetric cryptography (e.g.AES) or homomorphic encryption could be employed to respectively render communications safe or avoid the federator to be aware of data.
In the case of ``noise encryption'' we assume labels are randomly shuffled to avoid the emergence of unexpected correlations which may lead to trivial signal processing based attacks. The pre-shared seed allows to easily remove the noise. 
We present the algorithm in the sequence diagram in Figure~\ref{F:naive}. 
Since the matrices are shared upfront with the server, and from that point on all the calculations happen on the server, there is no need for iterative communication rounds to exchange data between the central server and clients. This very lightweight communication bandwidth requirement leads to a very fast execution, especially in scenarios with large datasets of many clients. 
The downside of this method is that an added security protection layer should be used, as the submatrices composing the kernel matrix $K$, containing distances (as we are using a RBF-Gaussian kernel) between sample points and landmarks are sent from hospitals and omics centers to the central server, and are thus vulnerable to attacks. Here we are assuming that we want to protect not only samples vectors but conservatively also their reciprocal distances. In Section~\ref{SS:EDM} we discuss to what extent this is a vulnerable step. In fact, it is known that given a Euclidean Distance Matrix (EDM), reconstructing the original points configuration up to roto-translation is easily done with an eigenvalue decomposition of the matrix. However the distances contained in $K$ are partial, and are not distances of the points between themselves, but points against random landmarks. We hypothesize (and empirically confirm) that for a small enough number of landmarks, this reconstruction is complex enough to make the communications in the naive method safe enough. Also we will discuss later a further countermeasure to render EDMs not usable at all to recover distances.

\subsection{A secure federalized kernel least square learning algorithm}
\label{SS:securefedCG}
To cope with the potential security issue of the naive algorithm discussed in Section~\ref{SS:naive} in terms of distributing kernel matrices, we devise here an iterative algorithm which avoids communicating these partial matrices and which instead communicates only intermediate results. Regarding labels the same reasoning applies as per the naive algorithm.

We report  the pseudocode for this methodology in Algorithm~\ref{A:main}. This algorithm performs a series of federated learning cycles.
The key insight in the algorithm is that we can decompose matrix-matrix multiplication operations like $K diag(\alpha)$ by cycling over the clients (where $diag(\alpha)$ constructs a diagonal matrix given the vector $\alpha$). This can be again obtained as the kernel is feature-wise multiplicatively separable. Hence all the key cycles of the code exploit this features. Also we prescribe both the server and the clients to perform the same update operations on the vector $p$ of CG. 

Summarizing, first we initialize the gradient ($G$) in a distributed fashion hiding both labels and the kernel matrices. This initialized gradient, and hence $p$ ($p=-G$), is distributed to all clients. This concludes the initialization. The server will next coordinate the iteration over clients (hospitals and omics centers) to determine the $pK^TKp$ coefficient which is next distributed to all clients to perform the same CG iterations. Hence in this approach the CG iterations are performed synchronously by both the server and the clients, yet the server coordinates the computation of the calculation of $pK^TKp$. Below we detail the algorithm steps.

\subsubsection{Initialization}
\label{SS:initialization}
During initialization, client (hospital) number $i$ computes an aggregate vector $v =  K_{H_i}\alpha - y + \eta$, where $K_{H_i}$ is the local contribution to the kernel matrix, $\alpha$ is the starting approximate solution (initialized randomly with the same seed for every client). The aim of this vector is to share with the server labels $y$ and local contributions to the kernel matrix.
At the end of the initialization, the server can compute the starting gradient $K^T K \alpha - K y$ without the regularization term $\lambda$ that, for ease of calculation is going to be added at a later stage, see~\eqref{E:NKRLSsystem}.

\paragraph{The function \textsc{FederatedGradient} }
\label{P:fedegrad}
Clients compute their aggregate vectors $v$ cycling on the omics centers who own the different features, using the \textsc{FederatedGradient} function, see Algorithm \ref{A:featureFed}. This is the engine handling vertical federation by leveraging the fact that the local contributions can be aggregated with matrix-matrix Hadamard products, and then they can be communicated together to the server. Again we leverage the peculiarity of kernel matrices to support the correctness of the computations. 
For simplicity of representation and without loss of generality, in the pseudocode we assume that each omics center owns a single feature. Note that \textsc{FederatedGradient} also outputs a quantity $K^T Kp_h$, that remains unused during initialization as the starting conjugate direction $p$ is initialized to 0. It will be used in the following calls of this function during iterations.

\paragraph{The aggregate vector $v$}
\label{P:noisy_v}
The aggregate vector $v$ represents a local residual error, and its scope is to communicate to the server labels $y$ and local contributions to the kernel matrix. In general we would assume that the server can obtain the labels from all the clients beforehand (and in certain cases it might be even the generator of some labels, for instance when the federating server is itself an omics center or a hospital), however in this example we show a possible way to mask them with a noise term $\eta$. Another possible way to safely share this value would be to use an OTP, or to rely on the security protocols of frameworks such as such as NVIDIA FLARE~\cite{NVFLARE}.

\paragraph{Noisy gradient}
The starting gradient $G=K^T K \alpha - K^T y$  gets aggregated by the clients at line 40, where the clients cycles on the kernel contributions $K_o$ received from the different omics center. It will be aggregated globally by the server at line 53.

\paragraph{On adding noise and denoising}
As mentioned when describing the aggregate vector $v$, adding a noise term is just one possible way to mask the communication of labels and intermediate values between clients and server. At lines 42 -- 51, clients remove the noise from the starting gradient, who is then communicated to the server, that adds the regularization term (line 54).

\subsubsection{Conjugate Gradient Iterations}
During iterations, we very much follow the classical CG method, with two nested loops per CG iteration; each hospital ($h$ index) loops over omics centers (internal loop). These computes local gradient contributions through the function \textsc{FederatedGradient}, discussed in Subsection~\ref{SS:initialization}.
Outside these loops we have the main loop which iterates over epochs (the CG main iterations).

\paragraph{Update step size and residual}
Updates of step size, conjugate directions and residual happen on the server as per the classical CG method. 
Same for error and stopping conditions evaluations. Remark that here we chose the residual's squared sum as error to check convergence as usual in CG.

\begin{figure}[H]
    \centering
    \includegraphics[width=.8\linewidth]{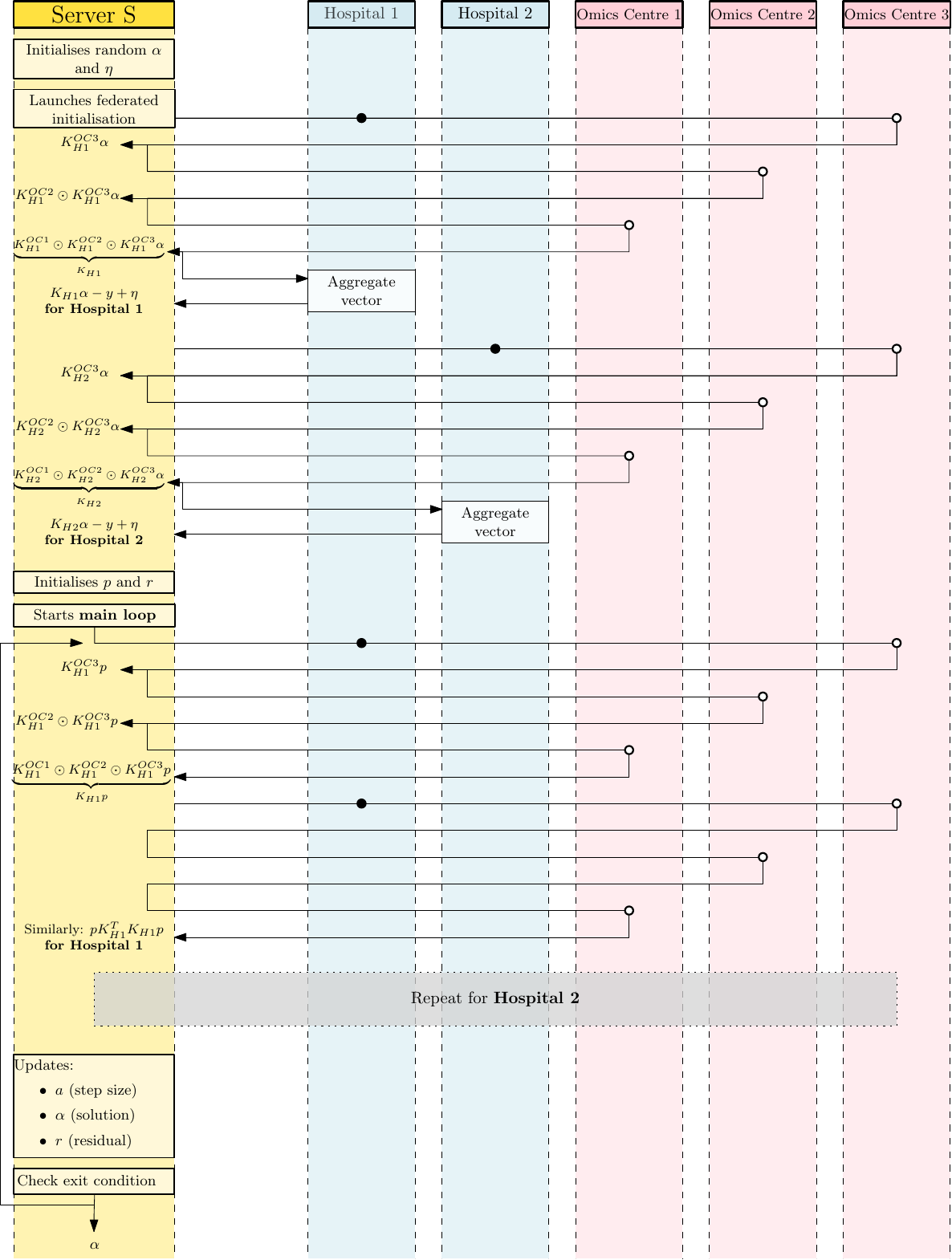}
    \caption{Sequence diagram for an instance of Algorithm~\ref{A:main} where we have two hospitals and three omics centers. Both the secure initialization and the main loop do a cycle on hospitals, where clinical data of different batches of patients reside ( horizontal federation: these iterations are represented with the $\bullet$ symbol). Then for each hospital both initialization and main loop cycle on the omics centers, where different features for the considered patients are stored, calling Algorithm~\ref{A:featureFed} (vertical federation: these iterations are represented with the $\circ$ symbol). The initialization is secure because neither the kernel submatrices nor the labels get ever sent without being modified by some random quantity known only by the server.}
    \label{F:fedCG}
\end{figure}

For a sequence diagram for Algorithm~\ref{A:main}, see Figure~\ref{F:fedCG}. A Python implementation of the procedure is available at \url{https://gitlab.iit.it/hpc/hybrid-fed-krls}. 

\begin{small}
\begin{breakablealgorithm}
\caption{Federated Conjugate Gradient (FedCG)}
\label{A:main}
\begin{algorithmic}[1]
\Function{FederatedCycles}{Inputs}
    \State \textbf{Inputs:} 
    \State $N_h$: number of clients (here hospitals)
    \State $X_{1:N_h}$: list of clients' samples matrices
    \State $y_{1:N_h}$: list of clients' labels
    \State $X_{test}$: test samples matrix
    \State $y_{test}$: test samples labels
    \State $W$: Nystr{\"o}m landmarks matrix
    \State $n_{nys}$: number of Nystr{\"o}m landmarks 
    \State \text{ker\_par}: type of kernel to be used
    \State $\lambda$: regularization parameter
    \State $toll$: tolerance
    
    \State \textbf{Outputs:} 
    \State $\alpha$: solution
    
    \\\hrulefill
    
    \State \emph{\textbf{Setting Global Parameters}}
    \State $N_e :=$ max number of global epochs
    \State $n_{feat} :=$ number of features
    \State $\alpha := $ approx. solution, randomly initialized
    \State $\eta := $ noise, randomly initialized
    \State $G :=$ zeroes vector of length $n_{nys}$
                \\\hrulefill
    \State \emph{\textbf{Initialization}}

    \State \emph{Computing intermediate values}

    \For{$h$ in range($N_h$)} \Comment{\textbf{On clients}}

        \State  $\_, \ v_h :=$ FederatedGradient($X_h, y_h$)\footnote{See Algorithm~\ref{A:featureFed}; each hospital cycles on omics centers and accumulates the partial gradient.}
        \If{$h = 0$} \Comment{\textbf{On server}}
            \State $v := v_h$
        \Else
            \State $v := v + v_h$
        \EndIf
        \State Move to next client (hospital)
    \EndFor

    \State \emph{Aggregating noisy starting gradient}
    \For{$h$ in range($N_h$)} \Comment{\textbf{On clients}}
        \For{$o$ in range($N_o, 0, -1$)}
            \If{$o == N_o$} 
                \State $K^TK \alpha \_ K^T y :=  (K_o^T \cdot diag(\eta))$ 
            \Else
                \State  $K^T K \alpha \_ K^T y := K_o^T \odot K^T K\alpha \_ K^T y$
            \EndIf
            \State $K^TK \alpha \_ K^T y_{red} :=$  sum each row in $K^TK \alpha \_ K^T y$
            \State $G = G + K^TK \alpha \_ K^T y$
        \EndFor            
        \State Move to next client (hospital)
    \EndFor

    \State \emph{Denoising gradient}
    \For{$h$ in range($N_h$)} \Comment{\textbf{On clients}}
        \For{$o$ in range($N_o, 0, -1$)}
            \If{$o == N_o$} 
                \State $noise\_rem := K_o^T \cdot diag(v)$ 
            \Else
                \State  $noise\_rem := K_o^T \odot noise\_rem $
            \EndIf
            \State  \Comment{\textbf{On server}}
            \State $noise\_rem_{red} :=$  sum each row in $noise\_rem$
            \State $G = G - noise\_rem_{red}$
        \EndFor
        \State Move to next client (hospital)
    \EndFor

    \State \emph{Initialization of CG parameters}   
    \Comment{\textbf{On server\footnote{The server will then send $G$ to all clients.}}}
    \State $G = G + \lambda  (I_{n_{nys}} \cdot\alpha)$
    \State $p := -G$ \Comment{Conjugate direction}
    \State $r:= p$ \Comment{Residual}

    \\\hrulefill
    \State \emph{\textbf{Conjugate Gradient Iterations}}
    \For{$epoch$ in range($N_{epochs}$)} \Comment{\textbf{On server}}
        \For{$h$ in range($N_h$)} \Comment{\textbf{On clients}}
            \State \emph{\textbf{Local gradient contributions:}}
            \State  $K^TKp_h, \ \_$ := FederatedGradient($X_h, y_h$)
            \State $K^TKp := K^TKp + K^TKp_h$ 
            \State $pK^TKp := pK^TKp + \text{dot}(p^T, K^TKp_h)$ 
        \EndFor
        \State {Server distributes $pK^T Kp$ to all clients}
        \State \emph{\textbf{Clients and Server synchronous updates}}
        \State $a := \frac{\text{dot}(r^T, r)}{pK^TKp}$ \Comment{\textbf{On server \& clients}}
        \State $\alpha := \alpha + a \cdot p$ 
        \State $r_{old} := r.copy()$ 
        \State $r := r - a \cdot K^TKp$ 

        \State $err := \sum (r^2)$ \Comment{Error}
        \\
        \State \emph{\textbf{Check if error is below the tolerance}}
        \If{$err < toll$} \Comment{\textbf{On server}}
            \State Exit the loop
        \EndIf
        \\
        \State $\beta := \frac{\text{dot}(r^T, r)}{\text{dot}(rold^T, rold)}$ \Comment{Coeff. for next direction}
        \State $p := r + \beta \cdot p$ \Comment{Conj. direction vector update}
    \EndFor

\Return $\alpha$
\EndFunction
\end{algorithmic}
\end{breakablealgorithm}

\end{small}

In Algorithm~\ref{A:main} we can see how the \emph{horizontal} partitioning (samples/patients distributed in several clients) is handled, through the same principles one would use when dividing the computation in batches of samples and then aggregating the parameters. The \emph{vertical} partitioning of the features (in our case, among omics centers) is handled by the \textsc{FederatedGradient} algorithm, for which we show the pseudocode in Algorithm~\ref{A:featureFed}.

\vspace{1em} 
\begin{small}
\begin{breakablealgorithm}
\caption{Federated Gradient updates for partitioned features: manages the vertical federation within FedCG}
\label{A:featureFed}
\begin{algorithmic}[1]
\Function{FederatedGradient}{Inputs}
    \State \textbf{Inputs:} 
    
    $X_h$: current client's sample matrix

    $y_h$: current client's labels embedded in a matrix
    
    $W$: Nystr{\"o}m landmarks matrix

    ker\_par: type of kernel to be used

    $\lambda$: regularization parameter

    $\alpha$: temporary solution

    $p$: conjugate vectors
    
    $\eta$: random noise
    
    \State \textbf{Outputs:} 
    
    $K^TKp$: accumulates partial gradients
    
    $v$: accumulates $K \alpha$ - $y$ + $\eta$

            \\\hrulefill

    \State \emph{\textbf{Parameters}}
    \State $N_o := \text{number of features\footnote{In this example we are assuming there is one feature per omics center}}$

                \\\hrulefill

    \State{\emph{\textbf{Compute the vertically federated gradient}}}
    \For{$o$ in range($N_o, 0, -1$)} \Comment{\textbf{On client}}
        \State $W_o$ := slice of W
        \State $K_i := \text{buildKernel}(X_{h_{o}}, W_{o}, \text{ker\_par})$
        \If{$o == N_o$}
            \State $K\alpha := K_i \cdot \text{diag}(\alpha)$ 
            \State $Kp := K_i \cdot \text{diag}(p)$

        \Else
            \State $K\alpha := K\alpha \odot K_i$   \Comment{Hadamard products}
            \State $Kp := Kp \odot K_i$
        \EndIf
    \EndFor
    \State $Kp :=$  sum each row in $Kp$

    \State $Y_h :=$ matrix embedding of $y_h$
    \State $v := \text{sum}((K\alpha - Y_h + \eta), axis=1)$
            \\\hrulefill
    \State \emph{\textbf{Features federation backward}}
    \For{$o$ in range($N_o, 0, -1$)} \Comment{\textbf{On client}}
        \State $K_i := \text{buildKernel}(X_{h_{o}}, W_{o}, \text{ker\_par})$ 
        \If{$o == N_o$}
            \State $K^TKp := K_i^T \cdot \text{diag}(Kp)$
        \Else
            \State $K^TKp := K^TKp \odot K_i^T$ \Comment{Hadamard products}
        \EndIf
    \EndFor

    \State $K^TKp:=$  sum each row in $K^TKp$
    \State $K^TKp := K^TKp + \frac{\lambda}{Nh} \cdot p$ \Comment{Add regularization term}
    
\Return $K^TKp,\ v$
\EndFunction
\end{algorithmic}
\end{breakablealgorithm}
\vspace{1em}
\end{small}

\subsection{Convergence}
We observe that while the gradient is computed in batches corresponding to the clients, it is then aggregated without any element of stochasticity, bringing out method on the steps of classical CG. Then, we can rely on the same convergence guarantees that we have in classical CG.

\subsection{Security}
\label{SS:security}
The strategy of using Nystr{\"o}m-like landmarks allows hospitals and omics centers to compute their part of the kernel matrix without sharing data. In fact, in our design each client would have an encrypted communication channel through a federated framework (such as NVIDIA FLARE~\cite{NVFLARE}) through which they can receive a seed to generate the landmarks and noise. Alternatively, one could use tokens and synchronize the communications between server and client.

From initialization onwards, in Algorithm~\ref{A:main}, exchanges between clients and the server contain only non-sensitive information. The only step that requires care is protecting the labels; this has been discussed in Subsection~\ref{SS:initialization}. In the example shown, labels are protected by adding noise (in the naive algorithm), which is then analytically removed by the server. Noise removal is possible because the clients and server are synchronized in terms of random number generation. Hence, once tackled this issue with this or another technique, the algorithm is unconditionally secure.

Note that the naive learning algorithm we proposed before prescribes the kernel matrix to \emph{travel} not-encrypted in the Internet. This procedure is secure under the assumption that from the Nystr{\"o}m kernel matrix it is not possible to recover the original distance matrix of all the samples.  In the next section we will try to understand if this assumption is likely to be true. 

\section{Experimental results}

\subsection{Performance}
\label{SS:Performance}
\begin{figure}[H]
    \begin{subfigure}[htb]{.45\linewidth}
        \centering        
        \includegraphics[width=\linewidth]{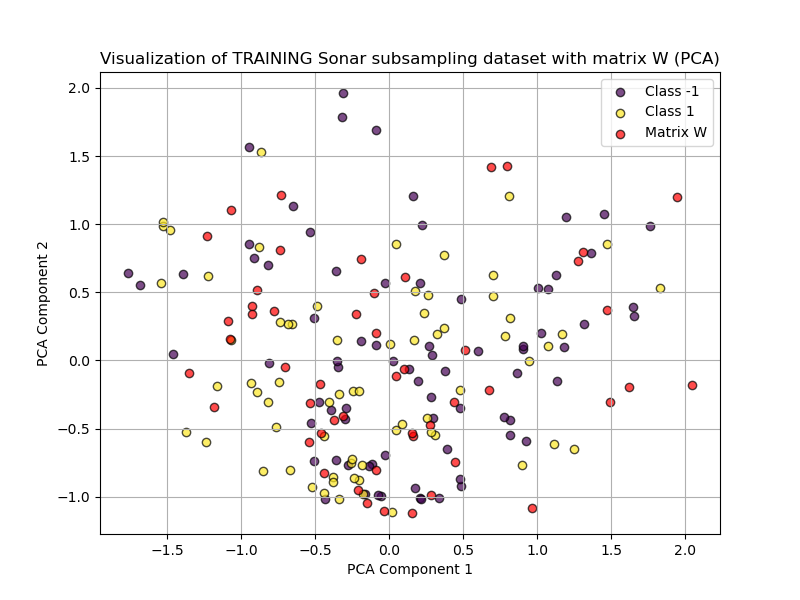}
        \caption{Training set of the Sonar dataset with its respective set $W$ of landmarks, selected with the $\mathcal{P}$ method.}
        \label{F:TrainWineSub}
    \end{subfigure}
    \begin{subfigure}[htb]{.4\linewidth}
        \centering        
        \includegraphics[width=.95\linewidth]{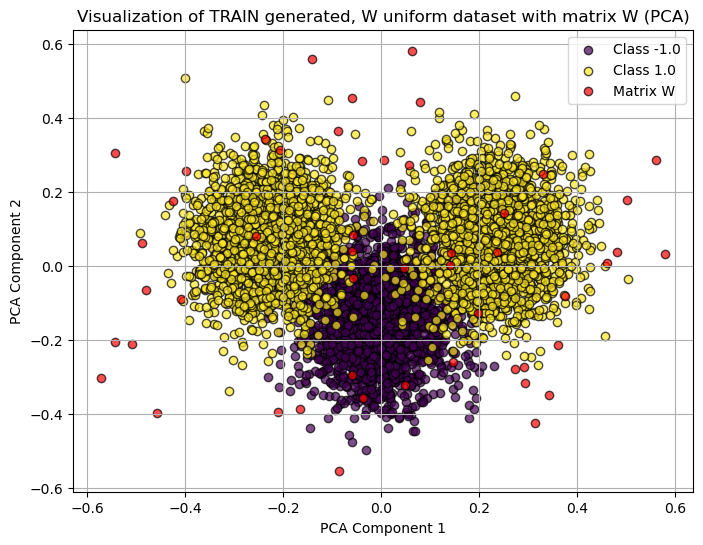}
        \caption{Training set of the Toy dataset with its respective set $W$ of landmarks, selected with the $\mathcal{U}$ method.}
        \label{F:TrainGenUni}
    \end{subfigure}
    \\
    \begin{subfigure}[htb]{.45\linewidth}
        \centering        
        \includegraphics[width=\linewidth]{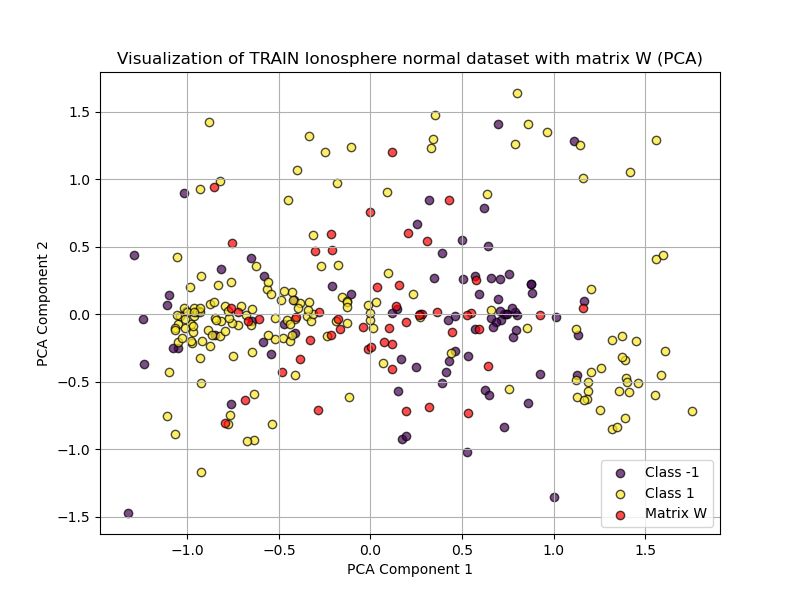}
        \caption{Training set of the Ionosphere dataset with its respective set $W$ of landmarks, selected with the $\mathcal{N}$ method.}
        \label{F:TrainBCNor}
    \end{subfigure}
    \caption{PCA visualizations of some of the training sets, showcasing the three different methods used to select the landmarks set $W$.}
\end{figure}

In this section we report performance results for our algorithms. We report the results of running Algorithm~\ref{A:main} (FedCG) and its the centralized version where there is no federation among hospitals. We call CenCG this centralized version which consists in centrally solving an RRLS problem with the CG method. CenCG is obtained by setting the number of hospitals $N_h$ to one in Algorithm~\ref{A:main}. We run them on different datasets, either generated or downloaded from the UCI Machine Learning Repository \url{https://archive.ics.uci.edu/}. 
The test performances reported have been run with 50 Nystr{\"o}m landmarks, generated either with the $\mathcal{P}$ method, the $\mathcal{U}$ method, or the $\mathcal{N}$ method. See Tables \ref{T:alpharandom_subsample}, \ref{T:alpharandom_uniform} and \ref{T:alpharandom_normal}  for simulations where the approximated solution $\alpha$ is initialized at random while the set of landmarks remains fixed. Moreover, in Table~\ref{T:datasets} we give a short description of the datasets, and in Figures~\ref{F:TrainWineSub}, \ref{F:TrainGenUni} and \ref{F:TrainBCNor} we show PCA visualizations of some of the training sets, together with the landmarks set $W$ selected with the three different sampling methods.

\begin{longtblr}[
  caption = {Datasets Descriptions},
  entry = {Datasets Descriptions},
  label = {T:datasets},
]{
  colspec = {X[-1]X[2]}, hlines,
  rowhead = 1, rowfoot = 0,
  row{1} = {font=\bfseries},
}
\textbf{Dataset} & \textbf{Description} \\
Toy dataset & Randomly generated training and test sets, each consisting of 6000 spatially distributed samples from three hospitals, with normalized features and a bias term added.\\
Fisher Iris  & \cite{iris}. Considered classes are: `setosa' and `versicolor or virginica' \\
Ionosphere  & \cite{ionophere}. Classes are `Good' and `Bad' (signals).\\
Sonar & \cite{sonar}. Classes are `Mine' and `Rock' \\
Breast Cancer Wisconsin & \cite{BCwisconsin}. Classes are `Benign' and `Malignant'. \\
Wine  & \cite{wine}. Classes are one type of wine against two others regrouped. \\
\end{longtblr}

\footnotesize
\begin{table*}
  \centering
  \caption{Performance  of FedCG and CenCG on various datasets. Metrics for random initialization. The matrix of Nystr{\"o}m points was built with the $\mathcal{P}$ method. Mean results for ten runs $\pm 2$ std.}
  \label{T:alpharandom_subsample}
  \begin{tabular}{
    @{}
    l
    l
    S[table-format=2.2]
    S[table-format=1.2]
    S[table-format=1.2]
    S[table-format=1.2]
    S[table-format=1.2]
    @{}
  }
    \toprule
    Dataset & Method & {Stop Epoch} & {Train Time} & {Test Accuracy} & {Test Recall} & {Test Precision} \\
   \midrule
        Toy dataset & FedCG & {$11.90 \pm 0.63$ }& {$3.29 \pm 1.88$} & {$0.94 \pm 0.00$} & {$0.96 \pm 0.00$} & {$0.95 \pm 0.00$ }\\
              & CenCG  &${11.90 \pm 0.63}$ & ${3.97 \pm 2.89}$ & ${0.94 \pm 0.00}$ & ${0.96 \pm 0.00}$ & ${0.95 \pm 0.00}$ \\
             \hline
    Fisher Iris & FedCG &${9.00 \pm 0.00}$ & ${0.23 \pm 0.13}$ & ${1.00 \pm 0.00}$ & ${1.00 \pm 0.00}$ & ${1.00 \pm 0.00}$ \\ 
                & CenCG  &${8.90 \pm 0.63}$ & ${0.04 \pm 0.04}$ & ${1.00 \pm 0.00}$ & ${1.00 \pm 0.00}$ & ${1.00 \pm 0.00}$  \\
              \hline
    Ionosphere & FedCG & ${139.30 \pm 2.99}$ & ${17.29 \pm 3.52}$ & ${0.92 \pm 0.00}$ & ${0.95 \pm 0.00}$ & ${0.88 \pm 0.00}$\\
              & CenCG  &  ${140.40 \pm 2.53}$ & ${5.19 \pm 3.11}$ & ${0.92 \pm 0.00}$ & ${0.95 \pm 0.00}$ & ${0.89 \pm 0.00}$ \\
              \hline
    Sonar     & FedCG &${46.80 \pm 1.26}$ & ${9.05 \pm 1.80}$ & ${0.89 \pm 0.00}$ & ${0.78 \pm 0.00}$ & ${0.91 \pm 0.00}$\\
            & CenCG  &${42.40 \pm 3.55}$ & ${1.76 \pm 0.33}$ & ${0.89 \pm 0.00}$ & ${0.78 \pm 0.00}$ & ${0.95 \pm 0.00}$\\
              \hline
    Breast Cancer & FedCG &${77.20 \pm 7.35}$ & ${8.63 \pm 2.92}$ & ${0.96 \pm 0.00}$ & ${0.95 \pm 0.00}$ & ${0.95 \pm 0.00}$\\
              & CenCG  & ${78.80 \pm 13.16}$ & ${2.70 \pm 1.34}$ & ${0.96 \pm 0.00}$ & ${0.95 \pm 0.00}$ & ${0.95 \pm 0.00}$\\
              \hline
    Wine      & FedCG & ${48.40 \pm 4.73}$ & ${2.12 \pm 0.41}$ & ${1.00 \pm 0.00}$ & ${1.00 \pm 0.00}$ & ${1.00 \pm 0.00}$\\
             & CenCG  & ${47.40 \pm 6.27}$ & ${0.37 \pm 0.12}$ & ${1.00 \pm 0.00}$ & ${1.00 \pm 0.00}$ & ${1.00 \pm 0.00}$ \\
    \bottomrule
  \end{tabular}
\end{table*}
\normalsize

\footnotesize
\begin{table*}[htb]
  \centering
  \caption{Performance of FedCG and CenCG on various datasets. Metrics for random initialization. The matrix of Nystr{\"o}m points was built with the $\mathcal{U}$ method. Mean results for ten runs $\pm 2$ std.}
  \label{T:alpharandom_uniform}
  \begin{tabular}{
    @{}
    l
    l
    S[table-format=2.2]
    S[table-format=1.2]
    S[table-format=1.2]
    S[table-format=1.2]
    S[table-format=1.2]
    @{}
  }
    \toprule
    Dataset & Method & {Stop Epoch} & {Train Time} & {Test Accuracy} & {Test Recall} & {Test Precision} \\
    \midrule
    Toy dataset & FedCG & ${16.80 \pm 0.84}$ & ${4.03 \pm 1.50}$ & ${0.94 \pm 0.00}$ & ${0.96 \pm 0.00}$ & ${0.95 \pm 0.00}$ \\
              & CenCG  & ${16.70 \pm 0.97}$ & ${4.63 \pm 3.30}$ & ${0.94 \pm 0.00}$ & ${0.96 \pm 0.00}$ & ${0.95 \pm 0.00}$ \\
              \hline
    Fisher Iris & FedCG & ${9.80 \pm 1.58}$ & ${0.21 \pm 0.06}$ & ${1.00 \pm 0.00}$ & ${1.00 \pm 0.00}$ & ${1.00 \pm 0.00}$ \\
              & CenCG  & ${10.10 \pm 1.99}$ & ${0.05 \pm 0.04}$ & ${1.00 \pm 0.00}$ & ${1.00 \pm 0.00}$ & ${1.00 \pm 0.00}$
 \\
              \hline
    Ionosphere & FedCG & ${60.00 \pm 4.52}$ & ${12.08 \pm 2.30}$ & ${0.89 \pm 0.00}$ & ${1.00 \pm 0.00}$ & ${0.84 \pm 0.00}$ \\
              & CenCG  & ${60.80 \pm 4.30}$ & ${2.52 \pm 0.87}$ & ${0.89 \pm 0.00}$ & ${1.00 \pm 0.00}$ & ${0.84 \pm 0.00}$
\\
              \hline
    Sonar     & FedCG &${48.90 \pm 0.63}$ & ${15.90 \pm 3.72}$ & ${0.90 \pm 0.00}$ & ${0.86 \pm 0.00}$ & ${0.86 \pm 0.00}$
 \\
              & CenCG  & ${49.10 \pm 0.63}$ & ${2.56 \pm 0.74}$ & ${0.90 \pm 0.00}$ & ${0.86 \pm 0.00}$ & ${0.95 \pm 0.00}$
 \\
              \hline
    Breast Cancer & FedCG &${42.30 \pm 1.35}$ & ${7.15 \pm 1.74}$ & ${0.94 \pm 0.00}$ & ${0.84 \pm 0.00}$ & ${0.98 \pm 0.00}$
 \\
              & CenCG  & ${42.20 \pm 3.86}$ & ${1.54 \pm 0.43}$ & ${0.94 \pm 0.00}$ & ${0.84 \pm 0.00}$ & ${0.98 \pm 0.00}$
 \\
              \hline
    Wine      & FedCG &${37.80 \pm 2.07}$ & ${2.78 \pm 0.83}$ & ${1.00 \pm 0.00}$ & ${1.00 \pm 0.00}$ & ${1.00 \pm 0.00}$
 \\
              & CenCG  &${37.10 \pm 2.57}$ & ${0.35 \pm 0.15}$ & ${1.00 \pm 0.00}$ & ${1.00 \pm 0.00}$ & ${1.00 \pm 0.00}$
 \\
    \bottomrule
  \end{tabular}
\end{table*}
\normalsize

\footnotesize
\begin{table*}
  \centering
  \caption{Performance  of FedCG and CenCG on various datasets. Metrics for random initialization. The matrix of Nystr{\"o}m was built with the $\mathcal{N}$ method. Mean results for ten runs $\pm 2$ std.}
  \label{T:alpharandom_normal}
  \begin{tabular}{
    @{}
    l
    l
    S[table-format=2.2]
    S[table-format=1.2]
    S[table-format=1.2]
    S[table-format=1.2]
    S[table-format=1.2]
    @{}
  }
    \toprule
    Dataset & Method & {Stop Epoch} & {Train Time} & {Test Accuracy} & {Test Recall} & {Test Precision} \\
   \midrule
        Toy dataset & FedCG & ${13.00 \pm 0.00}$ & ${7.74 \pm 3.27}$ & ${0.94 \pm 0.00}$ & ${0.96 \pm 0.00}$ & ${0.95 \pm 0.00}$
\\
              & CenCG  & ${13.00 \pm 0.00}$ & ${6.43 \pm 2.95}$ & ${0.94 \pm 0.00}$ & ${0.96 \pm 0.00}$ & ${0.95 \pm 0.00}$  \\
             \hline
    Fisher Iris & FedCG & ${11.20 \pm 0.84}$ & ${0.31 \pm 0.20}$ & ${1.00 \pm 0.00}$ & ${1.00 \pm 0.00}$ & ${1.00 \pm 0.00}$\\ 
                & CenCG  & ${11.10 \pm 1.75}$ & ${0.08 \pm 0.06}$ & ${1.00 \pm 0.00}$ & ${1.00 \pm 0.00}$ & ${1.00 \pm 0.00}$\\
              \hline
    Ionosphere & FedCG & ${66.30 \pm 3.89}$ & ${8.67 \pm 2.90}$ & ${0.85 \pm 0.00}$ & ${0.95 \pm 0.00}$ & ${0.84 \pm 0.00}$\\
              & CenCG  & ${67.00 \pm 3.77}$ & ${4.87 \pm 2.66}$ & ${0.85 \pm 0.00}$ & ${0.95 \pm 0.00}$ & ${0.82 \pm 0.00}$\\
              \hline
    Sonar     & FedCG & ${47.30 \pm 3.66}$ & ${7.51 \pm 1.34}$ & ${0.83 \pm 0.00}$ & ${0.85 \pm 0.00}$ & ${0.79 \pm 0.00}$ \\
            & CenCG  &${48.40 \pm 2.35}$ & ${3.07 \pm 1.18}$ & ${0.83 \pm 0.00}$ & ${0.85 \pm 0.00}$ & ${0.77 \pm 0.00}$\\
              \hline
    Breast Cancer & FedCG & ${55.80 \pm 4.09}$ & ${4.82 \pm 0.85}$ & ${0.95 \pm 0.00}$ & ${0.89 \pm 0.00}$ & ${0.98 \pm 0.00}$\\
              & CenCG  & ${53.70 \pm 7.00}$ & ${2.64 \pm 0.87}$ & ${0.95 \pm 0.00}$ & ${0.89 \pm 0.00}$ & ${0.98 \pm 0.01}$ \\
              \hline
    Wine      & FedCG & ${40.70 \pm 3.53}$ & ${1.48 \pm 0.17}$ & ${0.96 \pm 0.00}$ & ${0.91 \pm 0.00}$ & ${1.00 \pm 0.00}$\\
             & CenCG  &${41.00 \pm 3.77}$ & ${0.28 \pm 0.03}$ & ${0.96 \pm 0.00}$ & ${0.91 \pm 0.00}$ & ${1.00 \pm 0.00}$ \\
    \bottomrule
  \end{tabular}
\end{table*}
\normalsize

Since the two algorithms perform the same operations, to the point that the central method was run using the same code as the federated one, but setting the number of hospitals to~1, there was no need to perform any hyperparameters tuning to compare the federated and the central methods. One could carry out a hyperparameters tuning to improve performance on each specific dataset, but this is beyond the scope of this paper. 

\begin{figure}[htb]
    \centering
    \begin{subfigure}[t]{0.4\linewidth}
        \includegraphics[width=\linewidth]{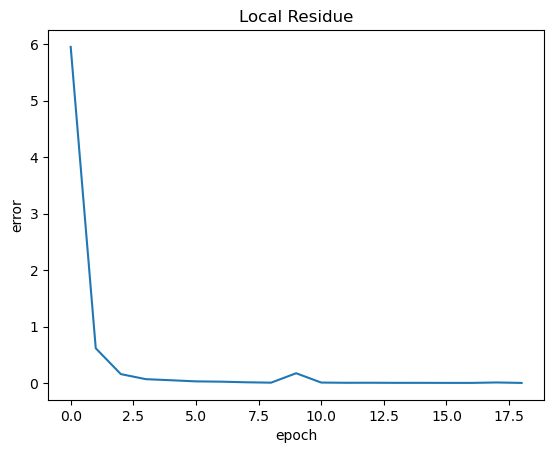}
        \caption{Local residue plot for FedCG on Sonar dataset.}
        \label{F:IrisFCG}
    \end{subfigure}
    \begin{subfigure}[t]{0.4\linewidth}
        \includegraphics[width=\linewidth]{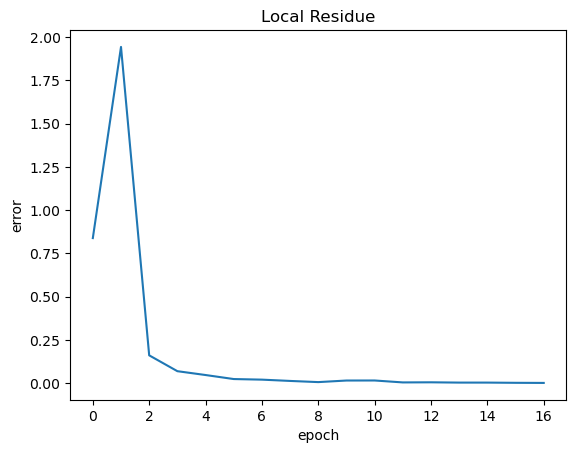}
        \caption{Local residue plot for CenCG on Sonar dataset.}
        \label{F:IrisCCG}
    \end{subfigure}
    \caption{Comparison of the local residue during training for FedCG and CenCG on the Sonar dataset}
\end{figure}

Both FedCG and CenCG converge to a solution within a similar number of iterations, based on their stopping epoch. This suggests that, in terms of convergence speed, they perform similarly. There doesn't seem to be a significant difference between their local residues, and in general we see a similar trend throughout the iterations, see Figures~\ref{F:IrisFCG} and~\ref{F:IrisCCG}.
FedCG tends to take slightly more computational time than CenCG for most datasets. This is expected because of the distribution of computations.

Results show that through FedCG we can obtain an exact version of CG while at the same time coping with a fully hybrid federation setting. 

\begin{figure}[htb]
    \centering
    \includegraphics[width=.6\linewidth]{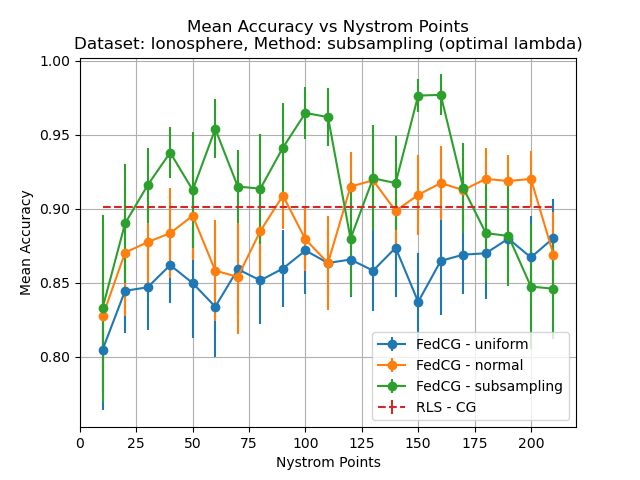}
    \caption{Mean accuracy (30 runs) for FedCG with the three different sampling methods $\mathcal{P}$, $\mathcal{U}$ and $\mathcal{N}$ (in the legend respectively: \emph{FedCG - subsampling, FedCG - uniform} and \emph{FedCG - normal}, at different levels of sampling (number of landmarks), compared with mean accuracy for the KRLS method solved iteratively with the CG method (in the legend \emph{RLS - CG}).}
    \label{fig:nystrom_regul}
\end{figure}

On top of these results we also briefly investigated 
the opportunity of employing this randomized version of KRLS in terms of its generalization capability, irrespective of its federated version. Indeed in~\cite[Theorem~1]{rudi_less_2015} it is proven that the number of Nystr{\"o}m landmarks,
selected with the classical $\mathcal{P}$ method, play the role of a regularization parameter. The plot in Figure~\ref{fig:nystrom_regul} suggests empirical evidence that selecting landmarks from other distributions, as we did with sampling methods $\mathcal{U}$ and $\mathcal{N}$, achieve comparable or better results.

Thus we conclude that our randomized approach not only well suits the technical needs of federated learning but also provides, as a bonus, regularization properties which can further enhance the generalization ability of the network.

\section{Complexity vs. security: the Euclidean Distance Matrix reconstruction problem}
\label{SS:EDM}
Both the naive algorithm proposed in Section~\ref{SS:naive} (see Figure~\ref{F:naive}) and the secure algorithm FedCG discussed in Section~\ref{SS:securefedCG} (see Algorithm~\ref{A:main}) are implementations of the CG method in a hybrid federated setting. We remark that the first one has a clear computational advantage on the second one, which has a higher computation overhead. However, the second one guarantees that no matrix containing distances of data points or labels are sent in isolation. Indeed we recall that in the naive algorithm by simply taking minus the logarithm of the kernel matrix one can recover the distance matrix and this poses a potential risk.
There are some mitigation points to the security risks posed by the naive algorithm: we remark that using Nystr{\"o}m-like landmarks, the distances within the (submatrices composing the) kernel matrix $K$ are not the Euclidean distances between points of the dataset, but Euclidean distances between points of the dataset and random landmarks. 
In particular, let us consider a set $X = \{x_1, \ldots, x_n\} \in \mathbb{R}^d$ of training samples, and a set $W=\{w_1, \ldots, w_m\}\in \mathbb{R}^d$ of landmarks, with $d, n, m \in \mathbb{N}$ and $m \leq n$. The matrix of shared distances would be: 
\[
D_{X, W} = \begin{bmatrix}
    \|x_1 - w_1\|^2 &   \cdots & \|x_1 - w_m\|^2 \\
    \|x_2 - w_1\|^2 &  \cdots & \|x_2 - w_m\|^2 \\
    \vdots& \ddots & \vdots \\
    \|x_n - w_1\|^2 & \cdots & \|x_n - w_m\|^2 \\
\end{bmatrix}\\
\]
where $\| \cdot \|$ represents the Euclidean distance. 

To assess the security of the naive algorithm, we need to understand how easy it would be to rebuild the whole distance matrix  $D_{X, X}$ if $D_{X, W}$ were to be leaked. 
The natural way to formalize this problem is by framing it as an Euclidean Distance Matrix (EDM) completion problem. 

An EDM is a matrix of squared Euclidean distances between points in a set, see~\cite{dokmanic_euclidean_2015} for a survey on the topic focused on computational aspects and applications and~\cite{Liberti:survey2014} for a geometrical stance. Once an EDM is known, it is easy to retrieve the original points configuration through eigenvalue decomposition of the matrix. EDM matrices are closely related to Gram matrices, and for construction admit several properties, such as having all 0 entries on the diagonal (and hence having a 0 trace), being symmetric and having all entries greater or equal to zero. A notable property that allows for computational techniques that we will see in the following to be feasible is that: the rank of an EDM corresponding to points in $\mathbb{R}^d$ is at most $d+2$, this completely independent from the number of points. 

In our case the problem could be formulated as follows: under which conditions (on the naive algorithm) the matrix 
\[
D_{X} = \begin{bmatrix}
    \|x_1 - x_1\|^2 &   \cdots & \|x_1 - x_n\|^2 \\
    \|x_2 - x_1\|^2 &  \cdots & \|x_2 - x_n\|^2 \\
    \vdots& \ddots & \vdots \\
    \|x_n - x_1\|^2 & \cdots & \|x_n - x_n\|^2 \\
\end{bmatrix}\\
\]
can be reconstructed, in case the matrix $D_{X, W}$ was to be leaked?

To test this we consider the following matrix: 
\[
D=
\left[\begin{array}{cc}
D_X & D_{X, W} \\
D_{X, W}^T & D_{W} \\
\end{array}\right]
\]
where $D_{W}$ is the EDM of the landmarks set $W$. Remark that in the case of the $\mathcal{U}$ and $\mathcal{N}$ methods, $X$ and $W$ are setwise distinct, while in the $\mathcal{P}$ method we have that $W \subset X$.

In an EDM completion problem, the question is: given a partially specified symmetric matrix $M$ with zero diagonal, determine the unspecified entries to make $M$ an EDM. In our case we will mask the $D_X$ block in $D$, and assess under what conditions it is possible to determine its entries. We rounded up four algorithms that have appeared in the recent literature to solve the problem. Three are specific to EDMs, while the fourth is a more general matrix completion algorithm.

We then set up an experiment where we try to retrieve $D_X$ for different numbers of landmarks.

\subsection{Selected algorithms and performance}
The five algorithms we rounded up from a literature review were: Alternating Descent \cite[Algorithm~3.1]{Parhizkar:2013}, \cite[Algorithm~4]{dokmanic_euclidean_2015}, an algorithm that is part of the multidimensional scaling techniques group; Rank Alternation \cite[Algorithm~2]{dokmanic_euclidean_2015}, an algorithm exploiting the rank property of EDMs that only leverages the observed, potentially noisy, distances and the embedding dimension of the point configuration; OptSpace~\cite{Keshavan:2010}, an algorithm designed especially for recovering a low-rank matrix from noisy data with missing entries; SDR \cite[Algorithm 5]{dokmanic_euclidean_2015} using rank-constrained semidefinite relaxation programming; and finally Soft-Impute \cite[Algorithm~1]{Mazumder:2010}, an algorithm using convex relaxation techniques to provide a sequence of regularized low-rank solutions for large-scale matrix completion problems. We used MATLAB implementations for the mention algorithms~\cite{dokmanicGithub}, and run the experiments on the IIT High Performance Computing Franklin machine\footnote{We run on CPU-only nodes which are equipped with 512 Gb of RAM.}.

After the first tests we excluded OptSpace and SDR because, while being from literature the most promising~\cite{Parhizkar:2013, dokmanic_euclidean_2015}, the time of computation was by far the slowest among the tested algorithms. In particular we had issues running SDR with the CVX interface for MATLAB, as it run out of memory when tested on dataset other than the Toy dataset. 

\begin{figure}[htbp]
    \centering
    \begin{subfigure}[b]{0.3\textwidth}
        \centering
        \includegraphics[width=\textwidth]{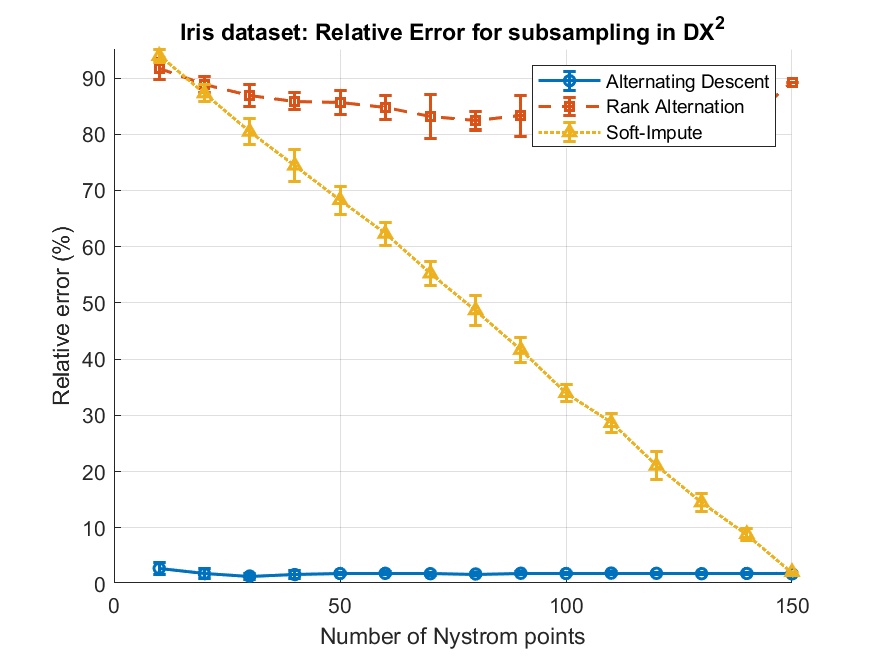}
        \caption{Iris dataset, landmarks generated with the $\mathcal{P}$ method.}
    \end{subfigure}
    \hfill
    \begin{subfigure}[b]{0.3\textwidth}
        \centering
        \includegraphics[width=\textwidth]{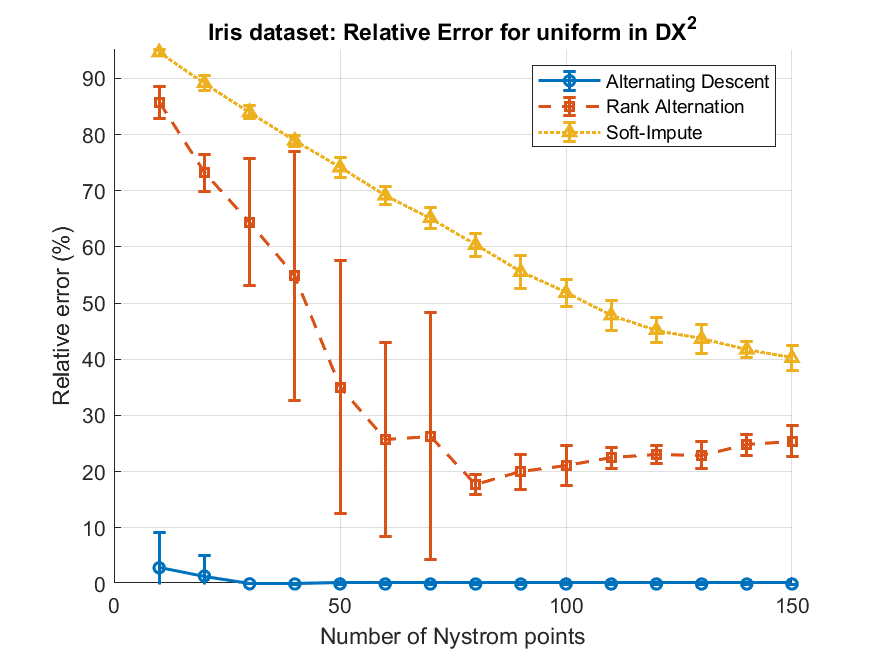}
        \caption{Iris dataset, landmarks generated with the $\mathcal{U}$ method.}
    \end{subfigure}
    \hfill
    \begin{subfigure}[b]{0.3\textwidth}
        \centering
        \includegraphics[width=\textwidth]{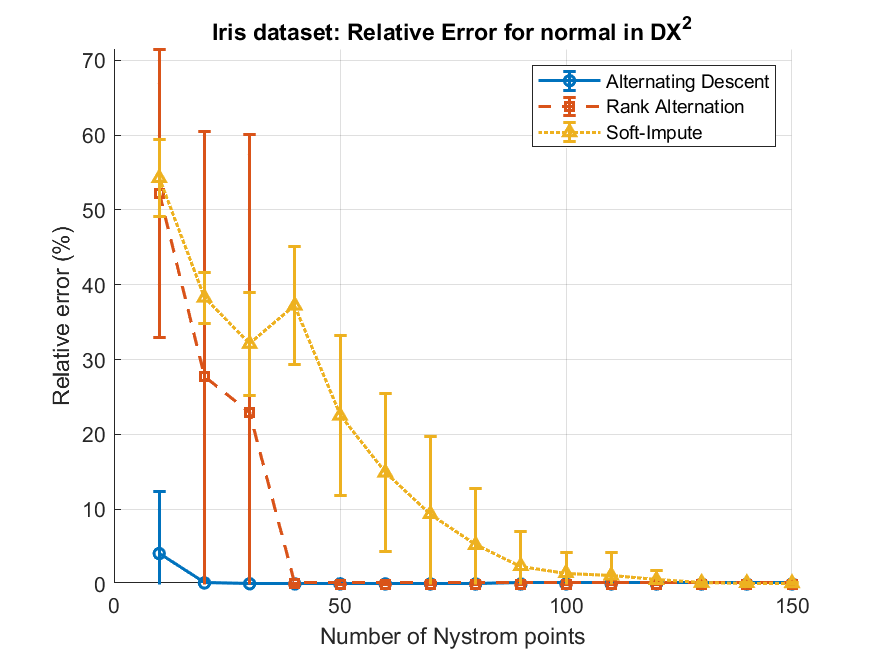}
        \caption{Iris dataset, landmarks generated with the $\mathcal{N}$ method.}
    \end{subfigure}
    
    \begin{subfigure}[b]{0.3\textwidth}
        \centering
        \includegraphics[width=\textwidth]{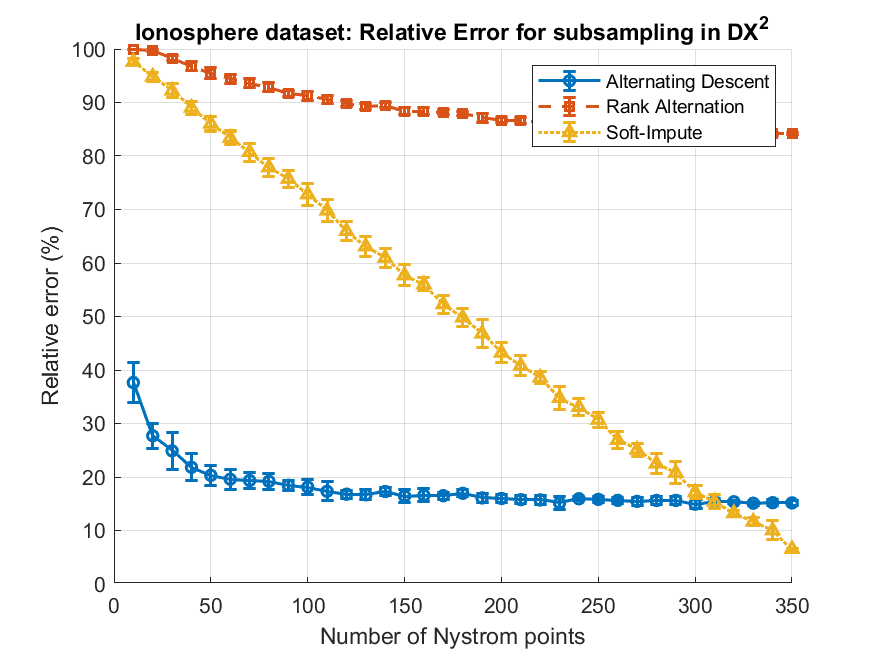}
        \caption{Ionosphere dataset, landmarks generated with the $\mathcal{P}$ method.}
    \end{subfigure}
    \hfill
    \begin{subfigure}[b]{0.3\textwidth}
        \centering
        \includegraphics[width=\textwidth]{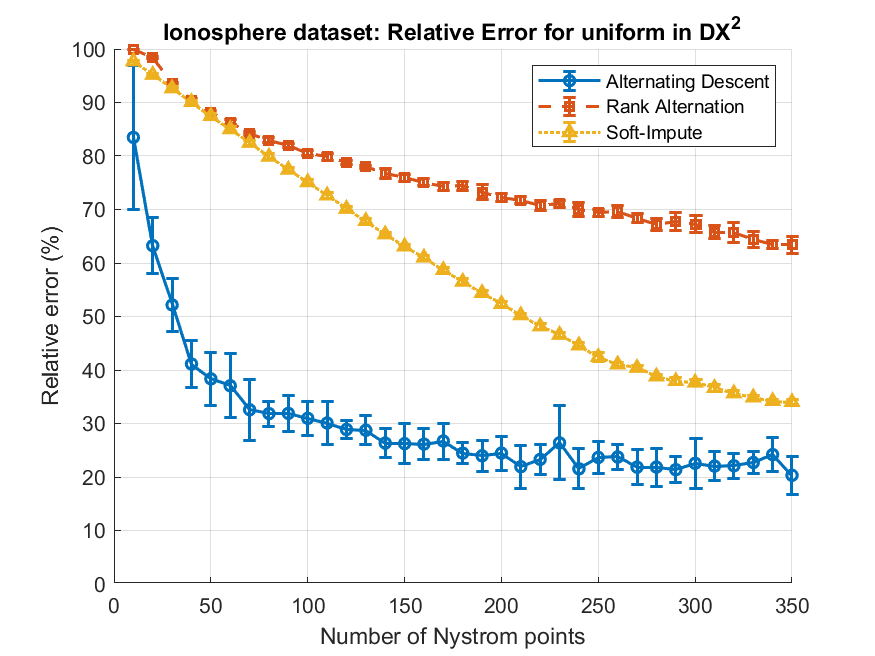}
        \caption{Ionosphere dataset, landmarks generated with the $\mathcal{U}$ method.}
    \end{subfigure}
    \hfill
    \begin{subfigure}[b]{0.3\textwidth}
        \centering
        \includegraphics[width=\textwidth]{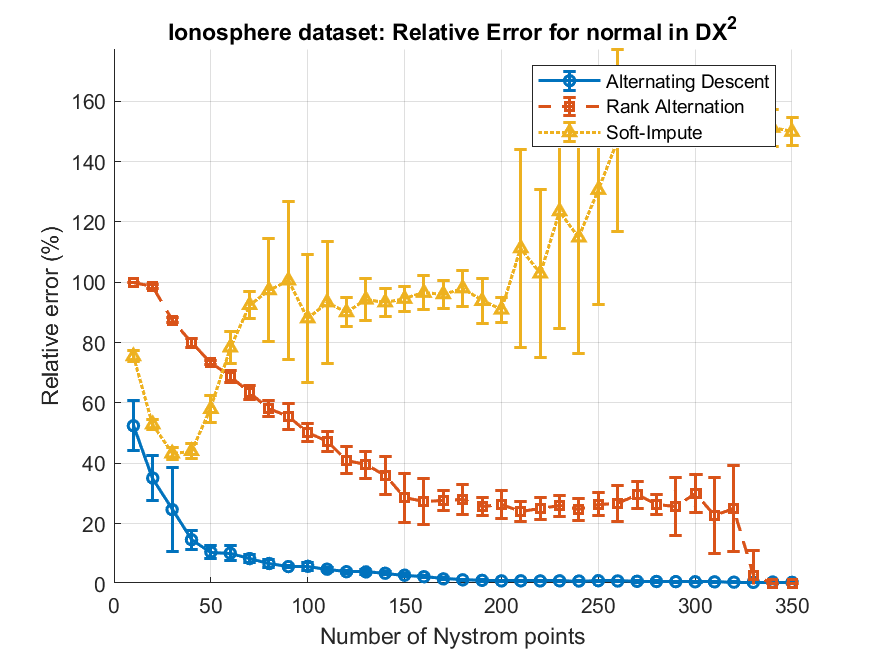}
        \caption{Ionosphere dataset, landmarks generated with the $\mathcal{N}$ method.}
    \end{subfigure}
    
    \begin{subfigure}[b]{0.3\textwidth}
        \centering
        \includegraphics[width=\textwidth]{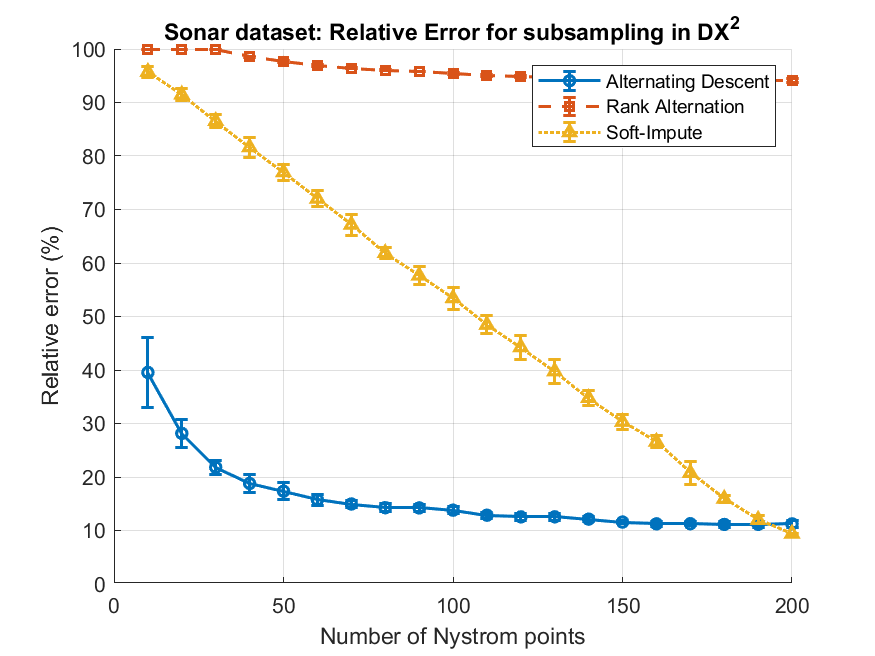}
        \caption{Sonar dataset, landmarks generated with the $\mathcal{P}$ method.}
    \end{subfigure}
    \hfill
    \begin{subfigure}[b]{0.3\textwidth}
        \centering
        \includegraphics[width=\textwidth]{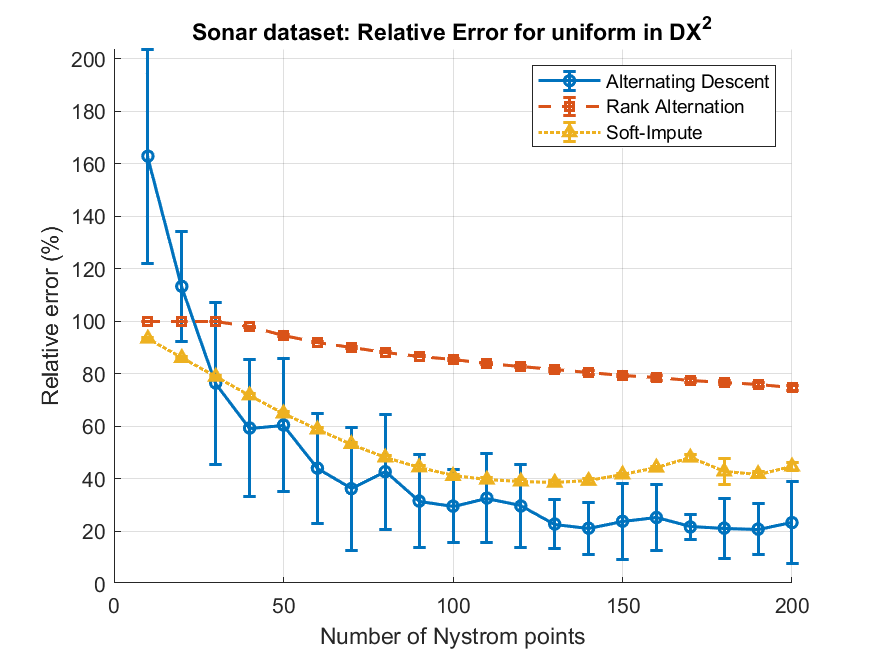}
        \caption{Sonar dataset, landmarks generated with the $\mathcal{U}$ method.}
    \end{subfigure}
    \hfill
    \begin{subfigure}[b]{0.3\textwidth}
        \centering
        \includegraphics[width=\textwidth]{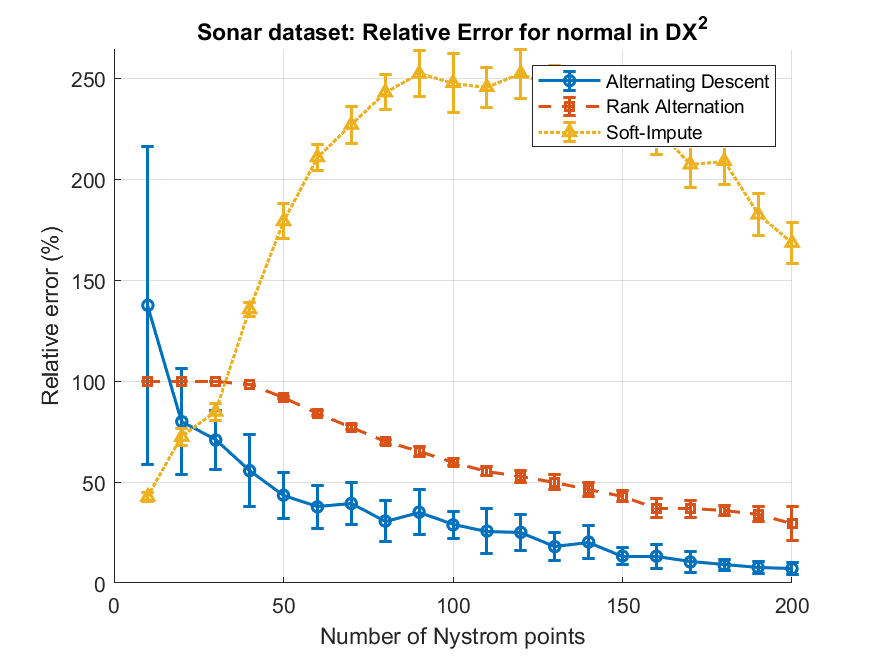}
        \caption{Sonar dataset, landmarks generated with the $\mathcal{N}$ method.}
    \end{subfigure}
    
    \caption{Plot of the relative error on the $D_X$ submatrix of $D$, for EDM completion algorithms on the Iris, Sonar and Ionosphere datasets.}
    \label{F:EDMreconstruction}
\end{figure}

We report the performance of Alternating Descend, Rank Alternation and Soft-Impute on the Iris, Ionosphere and Sonar datasets, for the three methods for generating the landmarks set $W$, in Figure~\ref{F:EDMreconstruction}. Results show that the Alternating Descent method is the most effective one, with Soft-Impute getting better results on the three tested dataset only for a high cardinality of $W$, and only for the $\mathcal{P}$ method for generating $W$. This is true also for the dataset with highest dimensionality (Ionosphere). With low dimensional dataset the reconstruction (e.g. Iris) the task appears more easily solved, whereas increasing dimensionality complicates the problem. We remark that the results depend on the data and landmarks distributions employed; we found that reconstruction methods are more challenged by the $\mathcal{U}$ and $\mathcal{N}$ landmarks distributions for $W$. Overall the $\mathcal{N}$ landmarks distribution seems to render the reconstruction problem slightly more complex to solve.  In general, increasing the number of landmarks favours the reconstruction, yet in most cases the benefit seems to saturate or is however limited. We recall that, thanks to regularization, the best model is often found with a reduced set of landmarks ~\cite{rudi_less_2015} (also see Figure \ref{fig:nystrom_regul}), this induces selecting smaller models, which at the same time mitigates significantly the risk of reconstruction. Indeed, a limited number of landmarks is beneficial both for generalization and in avoiding a too accurate reconstruction of the distance matrix from the tested algorithms.

\subsubsection{Randomization of kernel widths}
Considering that our aim is to use a kernel matrix and not specifically a distance matrix, a further countermeasure  we could employ to increase the security of naive FedCG is to
use random $\gamma$ parameters of the Gaussian kernel, a different one for each landmark.
That is the representation would become:
\begin{equation}
\label{E:representer_random_gamma}
f(x) = \sum_{j=1}^{m} \alpha_j \exp(-\gamma_j||x_j-x||)
\end{equation}
The problem would remain well defined, the network will be still generalizing~\cite{Gastaldo-Decherchi:2016} but the matrix completion methods specifically designed for EDMs would not be easily applicable anymore. This is because taking minus logarithm of the kernel matrix would not yield any distance matrix. Alternatively one could still use further EDMs methods able to cope with noisy observations, by looking at $\gamma_j d(x,x_j)$ as $(\gamma+\eta_j) d(x,x_j)$ where $\eta_j$ is a noise term; also in this case the reconstruction problem would become more complicated.

\section{Discussion and Conclusions}
In real-world situations features of the same patients are spread between hospitals and omics centers. Hospitals host clinical data, and omics centers are able to produce their omics counterpart of the same patients. While biosamples and labels (diagnosis) are most often shared between hospitals and omics centers, it would be advantageous having a hybrid learning scheme that prevents the explicit sharing of the experimental lab results (and clinical data). 
In this paper, we introduced hybrid federated variants of Kernel Regularized Least Squares that allow us to obtain the same solution as the original centralized algorithm. To make the algorithms secure, we propose several countermeasures. First, we employ randomization to avoid sharing the kernel landmarks points. Next, we use a reduced number of landmarks to hinder the reconstruction of the distance matrix, which we want to protect. This also has the potential advantage of improving the generalization ability and making the algorithm faster. In addition to this we also propose an iterative scheme which prevents sharing the kernel matrix. Lastly, we also suggest a further strategy that heavily limits the applicability of distance reconstruction algorithms.

A possible apparent limitation of the proposed scheme is that RRLS is a one layer convex method, unlike state-of-art deep learning methods. However, such method should be imagined as part of a more complex scheme we aim to develop. 
We envision employing deep learning methods on each omics layer separately (via regular federation) to build not only deep classifiers but also learn deep representations (last layer features). Regarding landmarks we can envision employing federated generative methods to produce them. That is a generative network is learnt for each omics in isolation to provide samples coming from the original distribution.
With these synthetic landmarks and the previously learnt deep representations, the hybrid RRLS method can be used to merge these representations in a single multi-omics classifier (or regressor). Hence, RRLS can be considered a tool acting as a final layer in a deep learning pipeline composed of several deeply learnt omics layer representations.

\section*{Acknowledgements}
This research was partially funded by the MISE project number F/180010/02/X43, \emph{Piattaforma Cloud Per la Diagnostica Avanzata Basata su Dati Clinico-Omici} and MISE Accordo per l'innovazione numero 96 \emph{Piattaforma per la Medicina di Precisione Intelligenza Artificiale e Diagnostica Clinica Integrata}. Authors acknowledge the support of the Data Science and Computation facility team for the HPC Franklin platform.

\appendix
\section{Supplementary Tables}

In Tables~\ref{T:nysrandom_subsample}, \ref{T:nysrandom_uniform} and \ref{T:nysrandom_normal} we report the performance of FedCG and CenCG in a simulation where the algorithm parameters were initialized only once at the beginning, but the matrix $W$ was newly generated at each run. 
\footnotesize
\begin{table*}[!ht]
  \centering
  \caption{Performance of FedCG and CenCG on various datasets. Metrics for simulations run with a fixed initialization and a newly generated matrix of Nystr{\"o}m points with the $\mathcal{P}$ method for each run. Mean results for ten runs $\pm 2$ std.}
  \label{T:nysrandom_subsample}
  \begin{tabular}{
    @{}
    l
    l
    S[table-format=2.2]
    S[table-format=1.2]
    S[table-format=1.2]
    S[table-format=1.2]
    S[table-format=1.2]
    @{}
  }
    \toprule
    Dataset & Method & {Stop Epoch} & {Train Time} & {Test Accuracy} & {Test Recall} & {Test Precision} \\
    \midrule
    Toy dataset & FedCG &${12.40 \pm 1.40}$ & ${3.56 \pm 2.88}$ & ${0.94 \pm 0.00}$ & ${0.96 \pm 0.01}$ & ${0.95 \pm 0.00}$ \\
              & CenCG  &${12.30 \pm 0.97}$ & ${3.16 \pm 2.75}$ & ${0.94 \pm 0.00}$ & ${0.96 \pm 0.01}$ & ${0.95 \pm 0.00}$ \\
              \hline
    Fisher Iris & FedCG &${9.00 \pm 0.00}$ & ${0.21 \pm 0.06}$ & ${1.00 \pm 0.00}$ & ${1.00 \pm 0.00}$ & ${1.00 \pm 0.00}$\\
              & CenCG  &${9.00 \pm 0.00}$ & ${0.04 \pm 0.02}$ & ${1.00 \pm 0.00}$ & ${1.00 \pm 0.00}$ & ${1.00 \pm 0.00}$
 \\
              \hline
    Ionosphere & FedCG &${136.30 \pm 23.89}$ & ${13.27 \pm 2.77}$ & ${0.89 \pm 0.04}$ & ${0.95 \pm 0.04}$ & ${0.87 \pm 0.06}$
  \\
              & CenCG  & ${140.70 \pm 29.14}$ & ${5.18 \pm 1.38}$ & ${0.89 \pm 0.04}$ & ${0.96 \pm 0.03}$ & ${0.87 \pm 0.06}$
\\
              \hline
    Sonar     & FedCG &${52.90 \pm 16.53}$ & ${8.76 \pm 2.91}$ & ${0.86 \pm 0.07}$ & ${0.83 \pm 0.15}$ & ${0.86 \pm 0.07}$
 \\
              & CenCG  &${53.30 \pm 18.38}$ & ${2.81 \pm 1.13}$ & ${0.86 \pm 0.07}$ & ${0.81 \pm 0.13}$ & ${0.87 \pm 0.06}$
\\
              \hline
    Breast Cancer & FedCG & ${102.90 \pm 23.50}$ & ${8.82 \pm 2.50}$ & ${0.97 \pm 0.02}$ & ${0.94 \pm 0.04}$ & ${0.97 \pm 0.05}$
\\
              & CenCG  & ${109.10 \pm 26.25}$ & ${4.02 \pm 1.33}$ & ${0.97 \pm 0.02}$ & ${0.94 \pm 0.03}$ & ${0.98 \pm 0.03}$
 \\
              \hline
    Wine      & FedCG &${77.60 \pm 21.06}$ & ${3.40 \pm 1.01}$ & ${0.98 \pm 0.03}$ & ${0.94 \pm 0.07}$ & ${1.00 \pm 0.00}$
 \\
              & CenCG &${80.30 \pm 18.67}$ & ${0.53 \pm 0.14}$ & ${0.98 \pm 0.03}$ & ${0.95 \pm 0.08}$ & ${1.00 \pm 0.03}$  \\
    \bottomrule
  \end{tabular}
\end{table*}
\normalsize
\footnotesize
\begin{table*}[!ht]
  \centering
  \caption{Performance of FedCG and CenCG on various datasets. Metrics for simulations run with a fixed initialization and a newly generated matrix of Nystr{\"o}m points, built with the $\mathcal{U}$ method, for each run. Mean results for ten runs $\pm 2$ std.}
  \label{T:nysrandom_uniform}
  \begin{tabular}{
    @{}
    l
    l
    S[table-format=2.2]
    S[table-format=1.2]
    S[table-format=1.2]
    S[table-format=1.2]
    S[table-format=1.2]
    @{}
  }
    \toprule
    Dataset & Method & {Stop Epoch} & {Train Time} & {Test Accuracy} & {Test Recall} & {Test Precision} \\
    \midrule
    Toy dataset & FedCG & ${18.60 \pm 5.51}$ & ${6.29 \pm 4.44}$ & ${0.94 \pm 0.00}$ & ${0.96 \pm 0.00}$ & ${0.95 \pm 0.00}$ \\
              & CenCG  & ${18.30 \pm 5.50}$ & ${4.86 \pm 2.59}$ & ${0.94 \pm 0.00}$ & ${0.96 \pm 0.00}$ & ${0.95 \pm 0.00}$ \\
              \hline
    Fisher Iris & FedCG & ${17.80 \pm 5.15}$ & ${0.37 \pm 0.13}$ & ${1.00 \pm 0.00}$ & ${1.00 \pm 0.00}$ & ${1.00 \pm 0.00}$\\
              & CenCG  &${18.40 \pm 4.24}$ & ${0.09 \pm 0.06}$ & ${1.00 \pm 0.00}$ & ${1.00 \pm 0.00}$ & ${1.00 \pm 0.00}$\\
              \hline
    Ionosphere & FedCG &${61.00 \pm 9.19}$ & ${6.30 \pm 1.54}$ & ${0.83 \pm 0.03}$ & ${0.99 \pm 0.05}$ & ${0.79 \pm 0.03}$\\
              & CenCG  & ${61.40 \pm 7.32}$ & ${2.31 \pm 1.02}$ & ${0.83 \pm 0.03}$ & ${0.99 \pm 0.04}$ & ${0.78 \pm 0.03}$ \\
              \hline
    Sonar     & FedCG &${27.30 \pm 5.74}$ & ${4.64 \pm 1.25}$ & ${0.71 \pm 0.07}$ & ${0.74 \pm 0.15}$ & ${0.62 \pm 0.10}$\\
              & CenCG  & ${26.20 \pm 5.23}$ & ${1.33 \pm 0.93}$ & ${0.71 \pm 0.07}$ & ${0.72 \pm 0.11}$ & ${0.64 \pm 0.07}$\\
              \hline
    Breast Cancer & FedCG &${59.80 \pm 15.34}$ & ${5.42 \pm 1.32}$ & ${0.94 \pm 0.03}$ & ${0.87 \pm 0.07}$ & ${0.96 \pm 0.04}$ \\
              & CenCG  & ${59.00 \pm 16.22}$ & ${2.42 \pm 1.60}$ & ${0.94 \pm 0.03}$ & ${0.86 \pm 0.09}$ & ${0.97 \pm 0.04}$\\
              \hline
    Wine      & FedCG &${64.60 \pm 12.90}$ & ${3.48 \pm 2.02}$ & ${0.96 \pm 0.05}$ & ${0.90 \pm 0.12}$ & ${0.99 \pm 0.03}$\\
              & CenCG  &${63.50 \pm 13.41}$ & ${0.53 \pm 0.30}$ & ${0.96 \pm 0.05}$ & ${0.92 \pm 0.11}$ & ${1.00 \pm 0.00}$  \\
    \bottomrule
  \end{tabular}
\end{table*}
\normalsize

\footnotesize
\begin{table*}[!ht]
  \centering
  \caption{Performance of FedCG and CenCG on various datasets. Metrics for simulations run with a fixed initialization and a newly generated matrix of Nystr{\"o}m points build with the  $\mathcal{N}$ method for each run. Mean results for ten runs $\pm 2$ std.}
  \label{T:nysrandom_normal}
  \begin{tabular}{
    @{}
    l
    l
    S[table-format=2.2]
    S[table-format=1.2]
    S[table-format=1.2]
    S[table-format=1.2]
    S[table-format=1.2]
    @{}
  }
    \toprule
    Dataset & Method & {Stop Epoch} & {Train Time} & {Test Accuracy} & {Test Recall} & {Test Precision} \\
    \midrule
    Toy dataset & FedCG &${12.60 \pm 1.93}$ & ${4.36 \pm 1.95}$ & ${0.94 \pm 0.00}$ & ${0.97 \pm 0.01}$ & ${0.95 \pm 0.00}$
 \\
              & CenCG  & ${13.30 \pm 4.12}$ & ${3.76 \pm 1.47}$ & ${0.94 \pm 0.00}$ & ${0.96 \pm 0.01}$ & ${0.95 \pm 0.00}$
 \\
              \hline
    Fisher Iris & FedCG & ${10.00 \pm 2.83}$ & ${0.20 \pm 0.07}$ & ${1.00 \pm 0.00}$ & ${1.00 \pm 0.00}$ & ${1.00 \pm 0.00}$
\\
              & CenCG  &${10.50 \pm 2.71}$ & ${0.04 \pm 0.02}$ & ${1.00 \pm 0.00}$ & ${1.00 \pm 0.00}$ & ${1.00 \pm 0.00}$
\\
              \hline
    Ionosphere & FedCG &${64.40 \pm 7.00}$ & ${6.99 \pm 1.92}$ & ${0.87 \pm 0.03}$ & ${0.99 \pm 0.02}$ & ${0.82 \pm 0.05}$
  \\
              & CenCG  & ${63.90 \pm 6.29}$ & ${2.60 \pm 1.44}$ & ${0.87 \pm 0.03}$ & ${0.99 \pm 0.02}$ & ${0.83 \pm 0.04}$
 \\
              \hline
    Sonar     & FedCG &${35.90 \pm 5.12}$ & ${6.18 \pm 1.55}$ & ${0.77 \pm 0.10}$ & ${0.68 \pm 0.16}$ & ${0.74 \pm 0.19}$
\\
              & CenCG  &${36.80 \pm 5.48}$ & ${1.89 \pm 1.02}$ & ${0.77 \pm 0.10}$ & ${0.67 \pm 0.13}$ & ${0.77 \pm 0.18}$
\\
              \hline
    Breast Cancer & FedCG & ${74.60 \pm 12.04}$ & ${6.84 \pm 0.97}$ & ${0.96 \pm 0.02}$ & ${0.92 \pm 0.04}$ & ${0.97 \pm 0.03}$
 \\
              & CenCG  &${76.00 \pm 11.31}$ & ${3.12 \pm 1.19}$ & ${0.96 \pm 0.02}$ & ${0.91 \pm 0.03}$ & ${0.98 \pm 0.04}$
 \\
              \hline
    Wine      & FedCG &${66.00 \pm 16.97}$ & ${2.68 \pm 0.52}$ & ${0.97 \pm 0.04}$ & ${0.92 \pm 0.14}$ & ${1.00 \pm 0.00}$
 \\
              & CenCG & ${65.40 \pm 15.12}$ & ${0.42 \pm 0.08}$ & ${0.97 \pm 0.04}$ & ${0.93 \pm 0.12}$ & ${1.00 \pm 0.00}$\\
    \bottomrule
  \end{tabular}
\end{table*}
\normalsize

\bibliography{main.bib}{}
\bibliographystyle{alpha}

\end{document}